\documentclass[journal]{IEEEtran}
\usepackage{amsmath,amsfonts}
\usepackage{amsthm,amssymb}
\usepackage{mathrsfs}

\usepackage{algorithm}
\usepackage{algpseudocode}  
\usepackage{array}
\usepackage[caption=false,font=normalsize,labelfont=sf,textfont=sf]{subfig}
\usepackage{textcomp}
\usepackage{stfloats}
\usepackage{url}
\usepackage{verbatim}
\usepackage{graphicx}
\usepackage{cite}

\usepackage{soul}
\usepackage{bm}
\usepackage{booktabs}
\usepackage{txfonts}
\usepackage{booktabs}
\usepackage{multirow}
\usepackage{caption3}
\usepackage{xcolor}

\usepackage{color}

\usepackage{listings}
\usepackage{pifont}
\usepackage{threeparttable}

\newcommand{\grad}{\nabla}

\newcommand{\bI}{\mathbf{I}}

\newcommand{\bzero}{\mathbf{0}}

\newcommand{\bx}{\mathbf{x}}

\newcommand{\bz}{\mathbf{z}}

\newcommand{\bepsilon}{{\boldsymbol{\epsilon}}}

\begin{document}

\title{Text2Earth: Unlocking Text-driven Remote Sensing Image Generation with a Global-Scale Dataset and a Foundation Model
% Text2Earth: Advancing Remote Sensing Text2Image Generation with a Global-Scale Dataset and the Foundation Model
}

% Chenyang Liu, ~\IEEEmembership{Graduate Student Member,~IEEE},
\author{Chenyang Liu, Keyan Chen, Rui Zhao, Zhengxia Zou,~\IEEEmembership{Senior Member,~IEEE}, \\and Zhenwei Shi$^*$,~\IEEEmembership{Senior Member,~IEEE}
% \\
% \vspace{4pt}
% Beihang University

\thanks{The work was supported by the National Natural Science Foundation of China under Grant 62125102, 624B2017, 62471014, U24B20177, and 623B2013, the National Key Research and Development Program of China under Grant 2022ZD0160401, the Beijing Natural Science Foundation under Grant JL23005, and the Fundamental Research Funds for the Central Universities. \emph{(Corresponding author: Zhenwei Shi (e-mail: shizhenwei@buaa.edu.cn))}
}
\thanks{Chenyang Liu, Keyan Chen, Zhengxia Zou and Zhenwei Shi are with the Department of Aerospace Intelligent Science and Technology, School of Astronautics, with the State Key Laboratory of Virtual Reality Technology and Systems, Beihang University, Beijing 100191, China, with the Key Laboratory of Spacecraft Design Optimization and Dynamic Simulation Technologies, Ministry of Education, and also with Shanghai Artificial Intelligence Laboratory, Shanghai 200232, China.

Chenyang Liu is also with Shen Yuan Honors College of Beihang University, Beijing 100191, China.

Rui Zhao is with the Department of Electrical and Computer Engineering, National University of Singapore, Singapore 117583.
}
}



\maketitle

\begin{abstract}
Recently, generative foundation models have significantly advanced large-scale text-driven natural image generation and have become a prominent research trend across various vertical domains. However, in the remote sensing field, there is still a lack of research on large-scale text-to-image (text2image) generation technology. Existing remote sensing image-text datasets are small in scale and confined to specific geographic areas and scene types. Besides, existing text2image methods have struggled to achieve global-scale, multi-resolution controllable, and unbounded image generation. To address these challenges, this paper presents two key contributions: the Git-10M dataset and the Text2Earth foundation model. Git-10M is a global-scale image-text dataset comprising 10.5 million image-text pairs, 5 times larger than the previous largest one. The dataset contains essential resolution information and covers a wide range of geographic scenes and contains essential geospatial metadata, significantly surpassing existing datasets in both size and diversity. Building on Git-10M, we propose Text2Earth, a 1.3 billion parameter generative foundation model based on the diffusion framework to model global-scale remote sensing scenes. Text2Earth integrates a resolution guidance mechanism, enabling users to specify image resolutions. A dynamic condition adaptation strategy is proposed for training and inference to improve image generation quality. Text2Earth not only excels in zero-shot text2image generation but also demonstrates robust generalization and flexibility across multiple tasks, including unbounded scene construction, image editing, and cross-modal image generation. This robust capability surpasses previous models restricted to the basic fixed size and limited scene types. On the previous text2image benchmark dataset, Text2Earth outperfoms previous models with a significant improvement of +26.23 FID and +20.95\% Zero-shot Cls-OA metric. Our project page is \emph{\url{https://chen-yang-liu.github.io/Text2Earth/}}

% We will publicly release the Git-10M dataset and the Text2Earth model at \emph{\url{https://chen-yang-liu.github.io/Text2Earth/}}

\end{abstract}


\section{Introduction} %Significance of the topic
\label{Significance}
Recently, generative foundation models have significantly advanced large-scale text-driven natural image generation and have become a prominent research trend across various vertical domains~\cite{li2024RSVLM_review,2023Generative_AI,zhang2024text2image_decade}, including medical imaging, autonomous driving, and virtual reality. These foundation models have demonstrated impressive image generation capabilities from large-scale image-text datasets, enabling them to produce large amounts of high-quality images. However, in the remote sensing field, there is still a lack of research on the large-scale text-to-image (text2image) generation technology based on foundation models~\cite{RS_foundation_model_survey,RS_foundation_model_survey_2,zhou2024towards_VLGFM,liu2024RSTVLM}.
This research holds considerable significance and application value, particularly in areas such as imaging simulation, virtual remote sensing scene construction, and data augmentation~\cite{Txt2Img_MHN,yu2024metaearth,map2rs}.

Unlike natural images, remote sensing images possess a unique ``God's-eye" perspective, characterized by wide geographical coverage, diverse scenes, and multiple resolutions~\cite{zhu2017_RSDL_review, 2024_AIRS_review,chen2023rsprompter,Liu_2022,Change_Agent}. These attributes underscore the necessity of the global-scale, multi-resolution controllable, and unbounded remote sensing text2image generation techniques. 

\begin{figure}
	\centering
 % \vspace{-15pt}
	\includegraphics[width=1\linewidth]{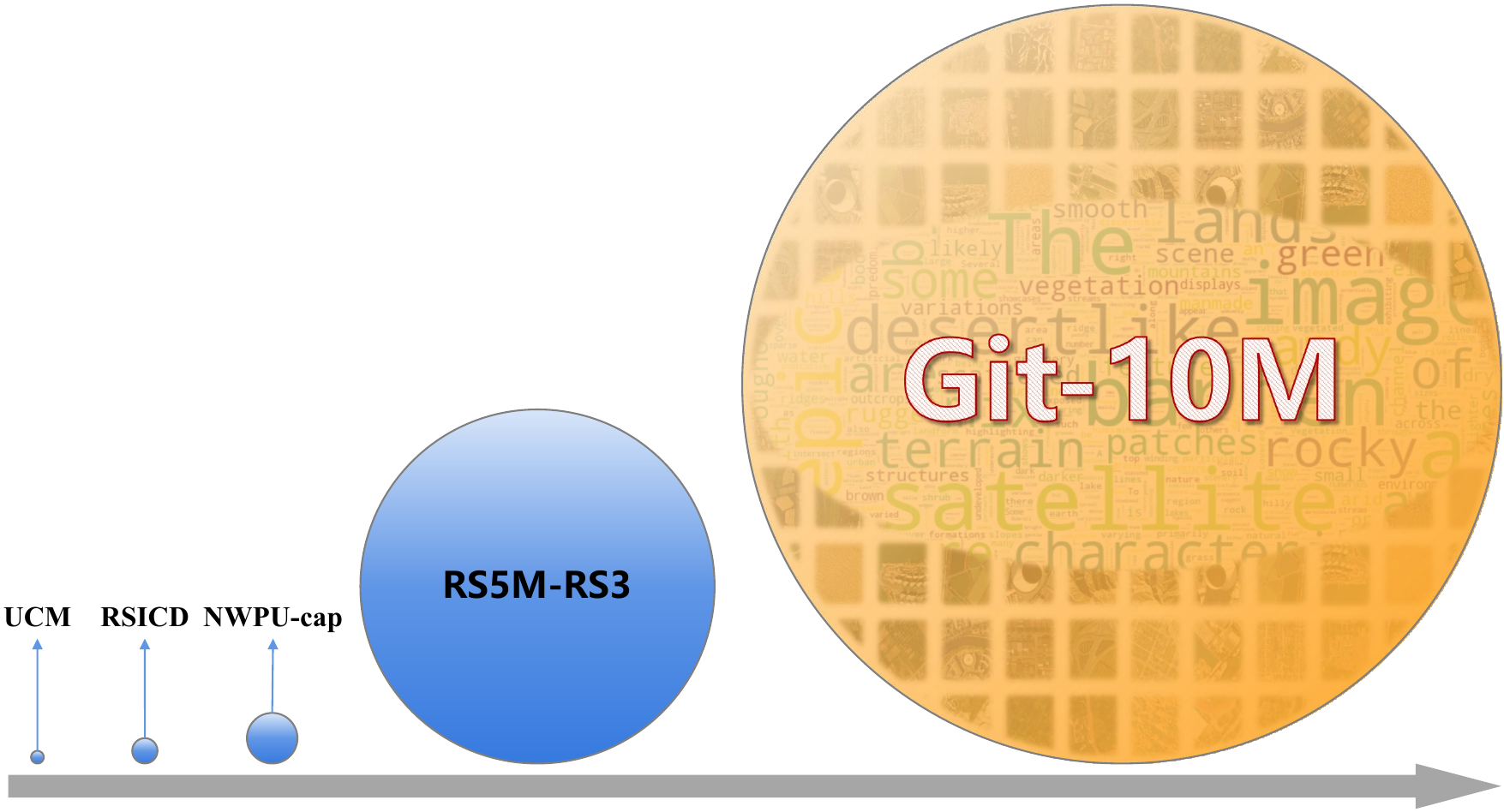}
        % \vspace{-10pt}
	\caption{Comparison between previous remote sensing text2image datasets and our Git-10M dataset.
 }
	\label{fig:fig_data_compar}
% \vspace{-15pt}
\end{figure}

Despite advancements in previous studies, significant challenges remain:
1)~\textbf{Dataset Limitations}: As illustrated in Fig.~\ref{fig:fig_data_compar} and Table~\ref{tab:dataset_cpmparison}, existing remote sensing image-text datasets are small-scale and lack sufficient diversity, such as UCM~\cite{qu2016deep_UCM_cap} and RSICD~\cite{Lu_2018_RSICD}. These datasets are typically confined to specific geographic areas and scene types. Moreover, these datasets usually consist of simple image-text pairs without crucial resolution information~\cite{Txt2Img_MHN}, restricting the flexibility of text2image generation in real-world scenarios that require images with specified resolutions.
2)~\textbf{Model Limitations}: 
Previous models have employed techniques like Generative Adversarial Networks (GANs) and Transformer to improve generation quality. However, these models struggle to adequately capture the complex structured geographical features inherent in global-scale remote sensing scenes. Meanwhile, they overlook the resolution-specific characteristics inherent in remote sensing imagery. This often results in the generation of images with uncertain resolutions, rather than tailored to user-specified needs.
Moreover, these models are restricted to basic fixed-size text2image generation, lacking the capability as foundation models to generalize across multiple text-driven generation tasks (e.g., unbounded scene construction and image editing), making them less versatile for real-world applications.

In this paper, we aim to advance remote sensing text2image generation towards global-scale scene generation, multi-resolution controllability, and unbounded large-size image synthesis through two primary contributions: a large-scale dataset and a powerful generative foundation model. 
To overcome the limitations of existing datasets, we developed the Git-10M dataset, a Global-scale image-text dataset comprising 10.5 million image-text pairs, which is 5 times larger than the previous largest dataset. Surpassing previous datasets, Git-10M encompasses a diverse range of global geographical scenes, including cities, forests, and mountains, while also containing rich metadata such as image resolution and geographic location. This comprehensive diversity empowers the model trained on Git-10M to generate realistic and global-scale images across various geographic scenes.

Building upon the Git-10M dataset, we propose Text2Earth, a 1.3 billion parameter generative foundation model based on the diffusion framework~\cite{dhariwal2021diffusion,StableDiffusion} to model global-scale remote sensing scenes. 
% Unlike traditional diffusion models that operate in pixel space, Text2Earth performs the diffusion in implicit space, significantly reducing the computational burden, particularly for large-scale remote sensing image generation. 
To efficiently generate large-scale remote sensing images, Text2Earth employs a VAE to compress images into a compact feature space. By performing the diffusion process in this feature space instead of conventional pixel space, Text2Earth significantly reduces computational overhead while preserving the image fidelity, making it well-suited for unbounded large-scale scene generation. To facilitate textual understanding, Text2Earth employ the OpenCLIP ViT-H encoder~\cite{CLIP} for robust and nuanced text representation, which is integrated into the denoising UNet network~\cite{Unet} via the cross-attention mechanism. We propose a resolution guidance mechanism for Text2Earth, addressing previous models' limitations in resolution control. Resolution-specific information is encoded and incorporated into each denoising step of the diffusion process, guiding noise prediction for resolution-controlled image generation. Furthermore, a dynamic condition adaptation strategy is proposed to integrate conditional inputs with null conditions to guide the denoising direction during training and inference. This strategy enhances generation quality while enabling the model to maintain performance in the absence of specific textual or resolution inputs.

The structure of the Text2Earth model equipped with 1.3 billion parameters. It consists of some core components: a Variational Autoencoder (VAE) for efficient image compression and reconstruction, an OpenCLIP ViT-H text encoder~\cite{CLIP} for converting text into high-dimensional semantic embeddings, a resolution embedding module, and a U-Net with the cross-attention mechanism for precise noise prediction.

Different from previous methods limited to generating fixed-size images with constrained scene diversity, Text2Earth not only supports resolution-controllable zero-shot text2image generation but also demonstrates robust generalization and flexibility across multiple tasks, including:
1) Zero-shot Text2Image Generation: Text2Earth can generate specific
image content based on user-free text input without requiring scene-specific fine-tuning. Additionally, on the previous remote sensing text2image benchmark dataset, Text2Earth surpasses prior models with a significant improvement of +26.23 FID and +20.95\% Zero-Shot Cls-OA metric.
2) Unbounded Remote Sensing Scene Construction: Text2Earth enables the unbounded generation of remote sensing scenes with a consistent spatial resolution through iterative user's text input, overcoming the fixed-size constraints of previous models. This functionality is ideal for creating expansive geographic visualizations.
3) Remote Sensing Image Editing: Text2Earth supports advanced editing tasks such as inpainting, cloud removal, and localized content modification, making it a versatile tool for interactive image editing.
4) Cross-modal Image Generation: Text2Earth has learned extensive knowledge and universal image generation capabilities from large-scale remote sensing data. These capabilities allow it for efficient transfer to diverse cross-modal image generation tasks, such as text-driven multi-modal image generation (e.g., NIR or SAR images), and image-to-image translation.
% Table \ref{tab:methods_cpmparison} provides a comparative analysis of Text2Earth and prior text2image models, highlighting its significant advancements.
% in terms of scalability, flexibility, and functionality.

\begin{table}%[htbp]%[!t]  %\small
\renewcommand{\arraystretch}{1.3}
\caption{Comparison between previous remote sensing text2image datasets and our Git-10M dataset.}
\label{tab:dataset_cpmparison}
\centering
% \resizebox{0.9\linewidth}{32mm}{
% \begin{tabular}{c |c c c c}
\centering
\begin{tabular}{m{60pt}<{\centering}|m{45pt}<{\centering}m{30pt}<{\centering}m{65pt}<{\centering}}
	\toprule%[1pt]
	%\hline
	Method & Global-Scale & Resolution & Number of Images\\ 
	\midrule
 	{UCM~\cite{qu2016deep_UCM_cap}} &\ding{55} &\ding{55} & 2.1k \\
        {RSICD~\cite{Lu_2018_RSICD}} &\ding{55} &\ding{55} & 10.9k \\
	{NWPU-cap~\cite{cheng2022nwpu_cap}} &\ding{55} &\ding{55} & 31.5k \\
 % {ChatEarthNet~\cite{yuan2024chatearthnet}} &\ding{55} &\ding{55} & 163k \\
 {RS5M-RS3~\cite{zhang2024rs5m}} & -- &\ding{55} & 2M \\
  % {SkyScript~\cite{wang2024skyscript}} &\ding{55} &\ding{55} & 2.6M \\ 这个数据集句子不流畅不适合text2image
      \midrule
	{Git-10M} &\ding{52} &\ding{52} & 10M \\
	\bottomrule
\end{tabular}
% }
\end{table}

% By addressing the gaps in the dataset and model, we believe Git-10M and Text2Earth offer a foundation that can drive future research in remote sensing image generation. 
Our contributions can be summarized as follows:
\begin{itemize}
\item 
\textbf{Global-Scale Dataset}: We present Git-10M, the largest-scale remote sensing image-text dataset, featuring extensive geographical diversity and metadata. It overcomes the limitations of previous small-scale datasets and provides a robust foundation for training generative models.
% overcoming the limitations of previous small-scale datasets by providing extensive geographical coverage and rich metadata.

\item 
\textbf{Generative Foundation Model}: We develop Text2Earth, a powerful diffusion-based generative foundation model that generates diverse global geographic scenes and multi-resolution images, ranging from close-up details to wide-area coverage, guided by user-provided textual input.
% Text2Earth is developed as a powerful foundation model for remote sensing text2image generation. It can generate diverse global geographic scenes and images of various resolutions, from detailed close-ups to wide-area coverage, based on the user's free text input.

\item 
\textbf{Generalization and Flexibility}: Text2Earth excels across various tasks, including zero-shot text2image generation, unbounded scene construction, image editing, and cross-modal image generation. This versatility represents a significant advancement, surpassing previous models restricted to fixed sizes and specific scenes. Besides, On the previous text2image benchmark RSICD dataset, Text2Earth surpasses the previous models with a significant improvement of +26.23 FID and +20.95\% Zero-shot Cls-OA metric. 
% This versatility makes it a valuable tool for a wide range of remote sensing applications, marking a significant advancement over earlier models that were limited to fixed-size images and specific scene types.
\end{itemize}

\section{Related Work}
In this section, we will review the recent advancements in generative foundation models and remote sensing text2image generation, highlighting the limitations of existing research.

\subsection{Generative Foundation Models in the Computer Vision}
% Foundation models have made significant progress in recent years. Models such as Stable Diffusion and DALLE have demonstrated remarkable improvements in both generalization ability and image quality~\cite{zhang2023text2image_review}, driven by their use of large-scale training data and the incorporation of innovative architectures like diffusion models and transformers. Compared with traditional models, a key advantage of foundation models is their strong transferability across various downstream tasks, such as image editing and style transfer, achieving excellent performance without requiring extensive data for fine-tuning. 

Generative foundation models (GFMs) have become increasingly influential in the field of computer vision, demonstrating remarkable advancements in the generation and transformation of visual data~\cite{zhang2024text2image_decade,zhang2023text2image_review,RSICCformer,liu2023decoupling,liu2022psnet}. These models, which are based on large-scale pre-training, are designed to capture a broad range of visual concepts and structures from vast datasets, making them versatile for numerous downstream tasks. Current generative models mainly focus on text2image generation. These models are typically built on three architectures: Generative Adversarial Networks (GANs), Autoregressive Transformers, and Diffusion models.

\subsubsection{GANs-Based Models}
GANs, introduced by Goodfellow \textit{et al.} in 2014~\cite{goodfellow2020_GAN}, are a classic generative model for text2image generation. In a typical GAN-based text2image framework, a generator learns to synthesize images from textual input, while a discriminator evaluates the realism of these images~\cite{tao2022df_GAN,esser2021taming_VQ_GAN,zhou2022towards_LAFITE,tao2023galip,ye2023recurrent,pan2023drag_GAN}. The adversarial interplay between these components fosters iterative refinement of generated images.
% This adversarial setup allows the generator to refine its output by competing with the discriminator.

Reed \textit{et al.} first used a conditional GAN (cGAN) structure~\cite{mirza2014conditional_GAN} to explore GAN-based text2image generation. StackGAN~\cite{zhang2017stackgan} generates high-resolution images in two stages: first by producing a low-resolution image from text, and then refining it to a high-resolution version. AttnGAN~\cite{xu2018attn_gan} introduced an attention mechanism that allowed the model to align specific words in the text with corresponding image regions. MirrorGAN~\cite{qiao2019mirrorgan} further emphasized bidirectional mapping between text and images to preserve textual coherence. In recent research, GigaGAN~\cite{kang2023scaling_GigaGAN} expands the model parameters and trains on large-scale data. It incorporates a multi-resolution hierarchical architecture and can generate ultra-high-resolution images at a faster speed. UFOGen~\cite{xu2024ufogen} combines GANs and diffusion models. It adopts a UNet architecture of the Stable Diffusion~\cite{StableDiffusion}, enabling it to leverage pre-trained Stable Diffusion for initialization, thereby significantly simplifying the training process. 
Despite these successes, GAN-based models are often hindered by challenges such as mode collapse and training instability, which can limit their effectiveness in generating diverse and high-quality images~\cite{gui2021review_GAN,jabbar2021survey_GAN}.

\subsubsection{Autoregressive Models}
Autoregressive models treat image generation as a sequential process. They typically leverage the large-scale Transformer architecture to generate images by sequentially predicting pixels or regions conditioned on preceding outputs and textual inputs~\cite{ramesh2021zero_DALLE,ding2021cogview,yu2022scaling_Parti,he2024mars,gafni2022make,liu2024rscama}. This approach has demonstrated strong capabilities in text2image generation by modeling the joint distribution of text and image tokens in a shared latent space.

OpenAI's DALL-E~\cite{ramesh2021zero_DALLE} laid the foundation for autoregressive text2image models with a two-stage training pipeline. It first trains a dVAE model to discretize the image, and then performs autoregressive modeling on the text and image tokens. Building on this, CogView~\cite{ding2021cogview} addresses the instability problem in large-scale autoregressive text2image training by proposing Precision Bottleneck Relaxation and Sandwich Layernorm. Different from decoder-only architecture, Parti~\cite{yu2022scaling_Parti} introduced an encoder-decoder architecture, treating text2image generation as a translation task, where the encoder processes text while the decoder predicts image tokens. Recent models emphasize efficiency and scalability. VAR~\cite{tian2024visual_VAR} proposes a coarse-to-fine ``next-scale prediction" mechanism, diverging from traditional ``next-token prediction". It achieved superior performance in terms of image quality, inference speed, and scalability compared to diffusion models. ZipAR~\cite{he2024zipar} accelerates autoregressive generation through a training-free parallel decoding framework, exploiting the spatial locality inherent in image data to enhance generation efficiency.

% These autoregressive models have the advantage of producing high-quality, coherent images, but they are computationally expensive due to their sequential nature. Their slow generation process often makes them less practical for real-time or large-scale applications.

\subsubsection{Diffusion-Based Models}
% Diffusion models represent a more recent advancement in generative modeling, and have shown great promise in high-quality image generation tasks, including text2image synthesis. Diffusion models operate by gradually adding noise to an image and then learning to reverse the process, effectively denoising the image to generate a final output. 
% This iterative denoising process allows the model to capture fine-grained details and produce images with greater diversity and realism compared to GANs and transformers.

Diffusion models have gained prominence as a leading approach in generative modeling~\cite{dhariwal2021diffusion,chen10654291_Spectral}. They operate by simulating a forward process that progressively corrupts data with noise and a reverse process that incrementally removes the noise, effectively reconstructing the original data~\cite{ho2020denoising}. This framework offers advantages such as training stability and the capacity to produce diverse, photorealistic images~\cite{zhang2023text2image_review}.

GLIDE~\cite{nichol2021glide} is a pioneering work comparing CLIP guidance and classifier-free guidance in text-conditional diffusion. DALL-E 2~\cite{ramesh2022_DALLE2} employs a two-stage approach that generates CLIP embeddings from textual descriptions and decodes these embeddings into detailed images. Stable Diffusion~\cite{StableDiffusion} introduces latent space diffusion for generating high-resolution images with reduced computational cost. Based on priors obtained from a large amount of data, it has become one of the most widely used generative foundation models and has enabled applications in domains such as artistic painting~\cite{rombach2022text,huang2022draw}, text-guided image editing~\cite{couairon2022diffedit,kawar2023imagic}, and text-to-video~\cite{singer2022makevideo,ho2022imagenvideo,pan2024synthesizing}. Recent innovations emphasize enhanced control and interactivity. ControlNet~\cite{zhang2023adding_ControlNet} enables spatial and structural control during image generation by integrating additional conditioning inputs. DragDiffusion~\cite{shi2024dragdiffusion} offers a point-based interface for precise spatial control, leveraging the power of pretrained diffusion models and latent space optimization at a single and carefully selected time step.

\subsection{Remote Sensing Text2Image Generation}
Remote sensing text2image generation task was first explored by Bejiga \textit{et al.}~\cite{bejiga2019retro_GAN}, who proposed a conditional GAN-based method to generate retro-images from ancient text descriptions of geographical landscapes. In subsequent works~\cite{bejiga2020retro,Bejiga2021_Improving}, they enhanced text encoding by using a doc2vec encoder~\cite{le2014distributed} to extract different levels of text information, such as object types, attributes, and spatial relationships. However, the generated images suffered from low resolution and insufficient detail, limiting their applicability. 
To address these issues, Zhao \textit{et al.}~\cite{zhao_StrucGAN} proposed StrucGAN, which generates high-resolution images through a multi-stage process. 
StrucGAN incorporates an unsupervised segmentation module within the discriminator to extract structural information from images, ensuring the synthesis of structurally coherent outputs. BTD-sGAN~\cite{chen2021remote_BTD_sGAN} introduced an innovative approach by replacing traditional Gaussian noise with Perlin noise and using segmentation masks and textual descriptions as conditional inputs to improve the quality of generated images.
% uses Perlin noise instead of traditional Gaussian noise, and takes segmentation masks and text as conditions to generate images.

Moving beyond GAN-based approaches, Xu \textit{et al.}~\cite{Txt2Img_MHN} developed Txt2Img-MHN, which employs a modern Hopfield network~\cite{ramsauer2020hopfield} to generate visual embeddings in an autoregressive manner. Their method leverages Vector Quantized Variational AutoEncoder (VQVAE)~\cite{van2017neural_VQVAE} and Vector Quantized Generative Adversarial Network (VQGAN)~\cite{esser2021taming_VQ_GAN} to discretize image embeddings. Additionally, Txt2Img-MHN implements coarse-to-fine hierarchical prototype learning for text and image embeddings via Hopfield Lookup, extracting representative prototypes from text-image embeddings.
% to enhance the coherence and quality of generated outputs.

Recent advancements have explored diffusion-based models for remote sensing text2image generation. Building on the Stable Diffusion model, DiffusionSat~\cite{txt2rs2_Diffusionsat} introduced a 3D ControlNet to extend the model’s capability for more conditional generation tasks. Similarly, CRS-Diff~\cite{txt2rs3_CRS_diff} also focus on controllable image generation. RSDiff~\cite{txt2rs4_rsdiff} adopts a two-stage text2image diffusion framework inspired by Imagen~\cite{saharia2022_IMAGEN}, where an initial low-resolution diffusion model generates preliminary images from textual inputs, followed by a super-resolution model that refines the images to achieve higher levels of detail.

% Different from previous autoregressive models using Transformer, Xu \textit{et al.}~\cite{Txt2Img_MHN} proposed Txt2Img-MHN by employing the modern Hopfield network~\cite{ramsauer2020hopfield} to generate visual embeddings in an autoregressive manner. They try Vector Quantized Variational AutoEncoder (VQVAE)~\cite{van2017neural_VQVAE} and the Vector Quantized Generative Adversarial Network (VQGAN)~\cite{esser2021taming_VQ_GAN} to obtain discretized image embeddings. Besides, Txt2Img-MHN performs coarse-to-fine hierarchical prototype learning on both text and image embeddings using a Hopfield Lookup and extracts the most representative prototypes from text-image embeddings. Recently, some studies have explored diffusion-based models. 
% Based on the Stable Diffusion, DiffusionSat~\cite{txt2rs2_Diffusionsat} designed a 3D ControlNet to extend more conditional generation tasks. RSDiff~\cite{txt2rs4_rsdiff} presents a two-stage diffusion model framework, which is similar to Imagen~\cite{saharia2022_IMAGEN}. It involves a low-resolution diffusion model that first creates initial images based on text inputs, followed by a super-resolution diffusion model which refines these images to achieve higher detail.

Despite the progress achieved by these models, significant challenges remain. Current approaches struggle to fully capture the complex and structured geographic features characteristic of global-scale remote sensing scenes, primarily due to the limited availability of diverse training datasets. This constraint limits their ability to generalize as foundation models for various text-driven generative tasks, such as unbounded scene construction and image editing.

% We conduct a comprehensive review of previous research at both the dataset and model levels. While earlier studies, including those based on GAN\cite{zhao_StrucGAN}, Transformer\cite{Txt2Img_MHN}, and Diffusion~\cite{txt2rs2_Diffusionsat,txt2rs3_CRS_diff,txt2rs4_rsdiff}, have advanced this technology, they are limited in terms of dataset scale, image generation quality, and resolution control. The datasets used in previous studies are small in size, with limited diversity in geographic scenes and lacking crucial resolution information. Previous methods failed to achieve controllable resolution, limiting their practical applications. 

\section{Global-scale image-text Dataset}
The Git-10M dataset is a global-scale remote sensing image-text pair dataset, consisting of 10.5 million image-text pairs with geographical locations and resolution information. This section will detail the construction process of the dataset and conduct a systematic analysis. 
% This section provides a detailed account of the dataset construction process, highlighting the innovative aspects, challenges faced, and the measures taken to ensure the dataset's high quality and utility for remote sensing text2image generation.

\subsection{Image Collection and Preprocessing}
\label{Image_Collection}
% The images are sourced from Google Earth and multiple publicly available datasets, such as Million-AID~\cite{million_aid} and SSL4EO-S12~\cite{SSL4EO_S12}. As shown in Fig. \ref{fig:dataset}, the dataset contains multiple resolutions from 0.5m/pixel to 128m/pixel and covers various geographical scenes (e.g., cities, forests, mountains, oceans, etc.). We implemented several quality assurance measures throughout the collection process, including removing redundant or low-quality images and balancing the dataset's geographic diversity. 

\begin{figure}
	\centering
 % \vspace{-15pt}
	\includegraphics[width=0.8\linewidth]{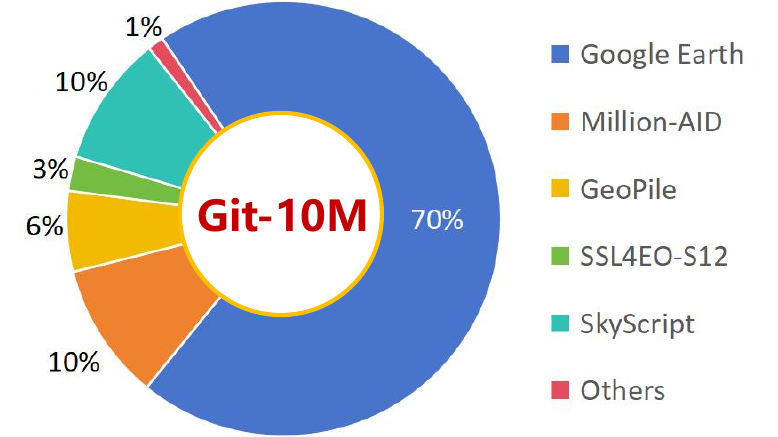}
        % \vspace{-10pt}
	\caption{The diverse image composition of the Git-10M dataset. Most images were collected from Google Earth, allowing public sharing and redistribution.
 }
	\label{fig:Source}
% \vspace{-15pt}
\end{figure}

As shown in Fig. \ref{fig:Source}, the images in the Git-10M dataset are sourced from multiple publicly available datasets and manually collected global remote sensing imagery from Google Earth. The public datasets, including Million-AID~\cite{million_aid}, GeoPile~\cite{GeoPile}, SSL4EO-S12~\cite{SSL4EO_S12}, SkyScript~\cite{wang2024skyscript}, DIOR~\cite{DIOR}, and RSICB~\cite{li2017rsiCB_dataset}, provide high-quality remote sensing images. These datasets primarily focus on scene classification tasks. During the collection process, we retained the scene category labels for each image to enable more precise semantic descriptions during the subsequent text annotation phase. These diverse data sources significantly enhance the richness of the Git-10M dataset.

To expand the dataset's scale and geographic coverage, we further collected remote sensing images with various resolutions and scene types from Google Earth. This collection process comprised two key steps: 1) randomly selecting regions worldwide to ensure broad sample distribution, and 2) manually selecting specific areas to ensure comprehensive coverage of typical geographic features such as urban areas, forests, mountains, and deserts. Throughout this process, we preserved metadata for each image, including geographic location and resolution, which provided essential support for subsequent analysis and text annotation.

After completing the image collection, we conducted stringent filtering and processing. First, duplicate or redundant ocean scenes were removed through manual screening to maintain diversity in geographic distribution. Additionally, a subset of images exhibited issues with visual quality, such as noise and artifact, which could negatively impact the training of image generation models. To address this, an image enhancement model was trained on a private high-quality remote sensing dataset and applied to all collected images, significantly improving the overall image quality of the dataset. {During the training of the model, we simulate various image degradation processes, such as blurring, noise addition, and compression, to create paired low-quality and high-quality images. We train the model using these paired images to learn the mapping from degraded images to their high-quality counterparts. This enhancement process helps to standardize image quality across the Git-10M dataset, making it more suitable for high-quality generative modeling. We will also release the enhancement model} at \emph{\url{https://github.com/Chen-Yang-Liu/Text2Earth}}

Through the above multi-stage collection and processing workflow, Git-10M not only achieves a breakthrough in scale, but also shows remarkable improvements in quality, diversity, and geographical coverage.

\begin{figure*}
	\centering
 % \vspace{-15pt}
	\includegraphics[width=1\linewidth]{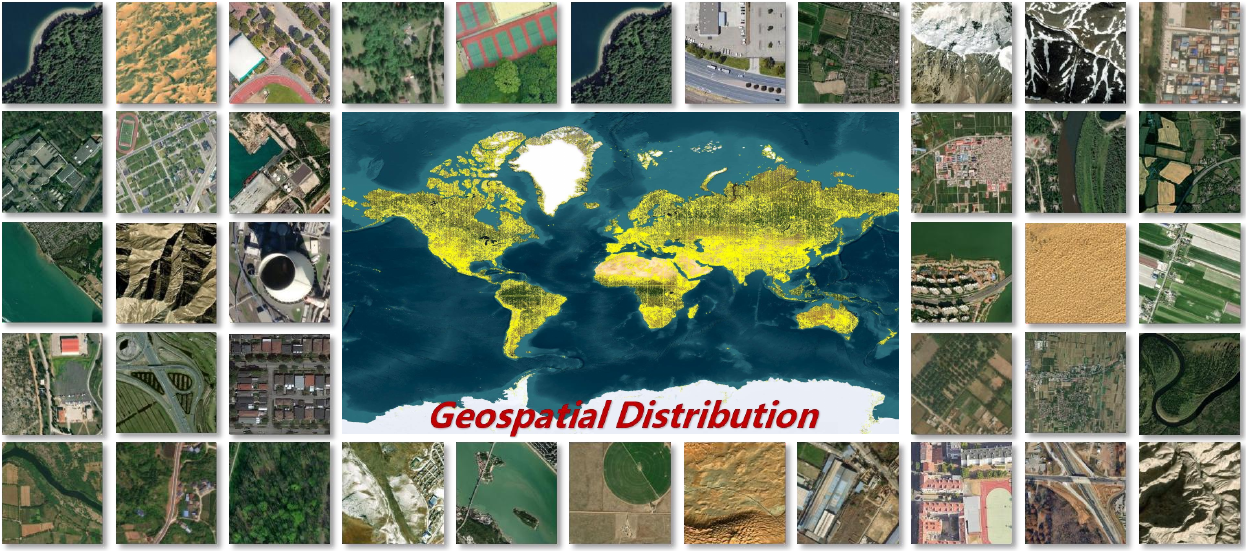}
        % \vspace{-10pt}
	\caption{The diverse geospatial distribution of the Git-10M dataset. {The yellow pixels represent the geographic locations where remote sensing images in Git-10M were sampled. The distribution shows that our dataset covers multiple continents and geographical regions, covering various typical scenes such as urban areas, forests, mountains, and deserts.}
 }
	\label{fig:Geo_distribution}
% \vspace{-15pt}
\end{figure*}

\subsection{Text Annotation}
% Given the large scale of the dataset, manual text annotation was impractical. To address this, we developed an automated annotation pipeline, which integrates OpenAI's GPT-4o model for generating text descriptions of the images. This pipeline incorporates error detection and correction mechanisms, ensuring high accuracy. We designed a dynamic prompting strategy that adjusts to image characteristics to improve the relevance of the generated texts. Additionally, we implemented an audit process with periodic sampling and human review to monitor and evaluate the annotation quality. The word cloud of all collected texts is shown in Fig. \ref{fig:cloud}.

Given the scale of over 10 million images, manual annotation of textual descriptions was infeasible. To address this challenge, we designed a automated annotation pipeline capable of efficiently generating high-quality text descriptions that accurately reflect image content. This pipeline leverages the GPT-4o~\cite{achiam2023gpt4_rep} API from OpenAI, combined with prompt optimization and annotation review strategies, to ensure both efficiency and accuracy.

For images with metadata such as geographic location, resolution, or scene category labels, these attributes were incorporated as additional context in the prompts provided to the GPT-4o model, significantly improving the relevance of the generated text. For example, when processing an image labeled as an airport scene, the scene information ``airport" was included in the prompt to guide the model toward generating a more semantically accurate description. To enhance the quality of text generation, the input prompts for GPT-4o underwent multiple iterative refinements. Compared to simple straightforward instructions like ``Describe the image content," we developed more sophisticated prompts emphasizing semantic details such as scene context and geographic features.

To ensure the reliability of the large-scale annotation, we established a review mechanism combining automated auditing and manual sampling inspections. The automated auditing process addressed potential issues arising from GPT-4o timeout responses or network errors, which could result in incorrect textual outputs due to unsuccessful image uploads. Additionally, periodic manual sampling was conducted to evaluate the accuracy of the generated text. Errors identified during the review process were fed back into the annotation pipeline, prompting prompt design refinements and reprocessing of erroneous samples.

This automated annotation pipeline successfully generated high-quality, semantically rich, and contextually accurate text descriptions for every image in the dataset. This provided critical support for the construction of Git-10M as a robust, high-quality resource for the remote sensing community. 

\subsection{Dataset Analysis}
% We conducted a comprehensive statistical analysis of both the images and corresponding textual annotations in the Git-10M dataset. This analysis covers several key aspects, including geographical coverage across diverse regions, the distribution of images from various data sources, the distribution of image resolutions, and image quality evaluations using quantitative metrics. In addition, we analyzed text features such as sentence length and complexity. This multi-dimensional analysis highlights the diversity and richness of Git-10M, establishing it as a robust training resource for advancing remote sensing text2image generation research.

To comprehensively evaluate the quality and diversity of the Git-10M dataset, we conducted a systematic analysis of both the image and corresponding textual annotations from multiple dimensions. The analysis includes the following aspects:

\begin{figure}
	\centering
 % \vspace{-15pt}
	\includegraphics[width=1\linewidth]{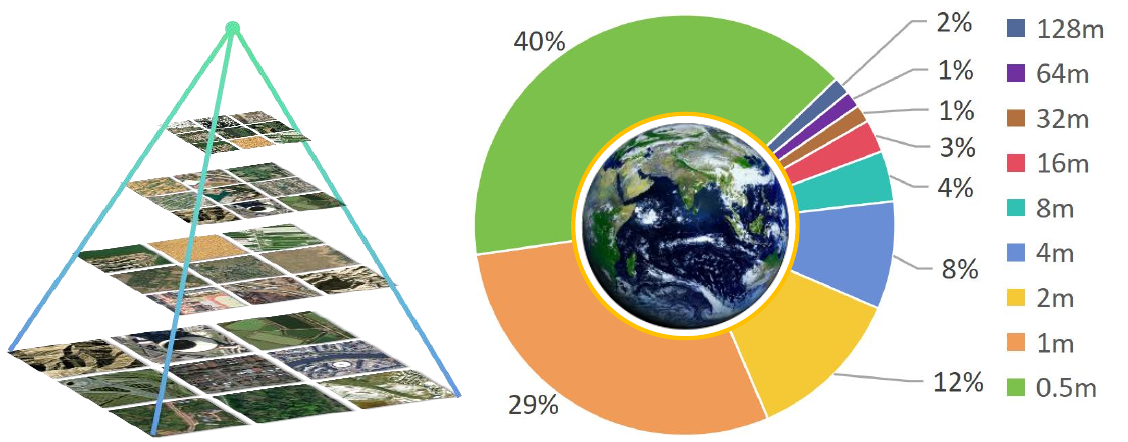}
        % \vspace{-10pt}
	\caption{The distribution of images with varying resolutions in the Git-10M dataset. The dataset encompasses images ranging from high resolution (e.g., 0.5m/pixel) to low resolution (e.g., 128m/pixel).
 }
	\label{fig:resolution}
% \vspace{-15pt}
\end{figure}

\begin{figure}
	\centering
 % \vspace{-15pt}
	\includegraphics[width=1\linewidth]{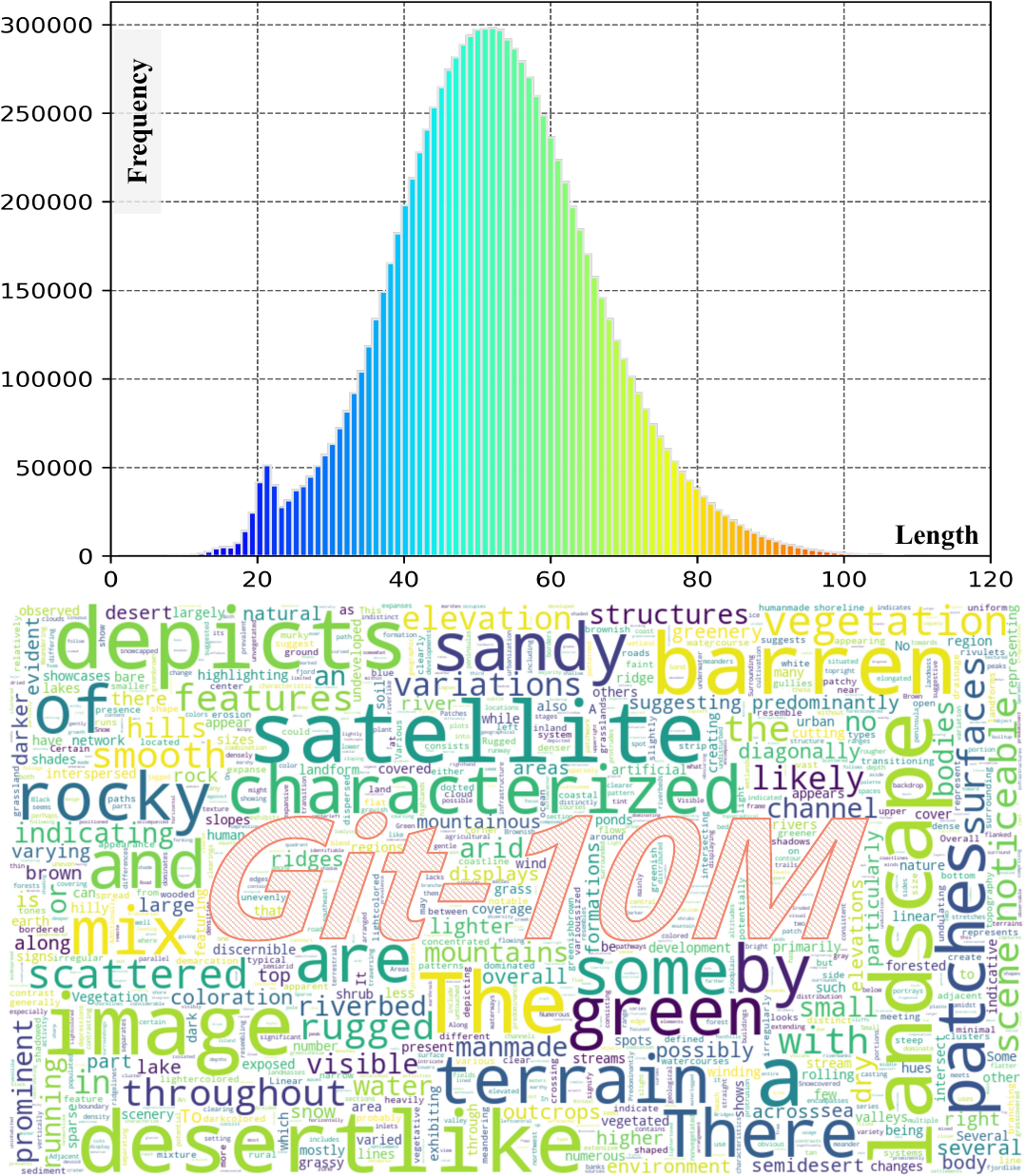}
        % \vspace{-10pt}
	\caption{{Text Analysis. Top: the word cloud of the texts in the Git-10M dataset. Bottom: the distribution of text lengths in the Git-10M dataset shows that each textual description averages approximately 52 words, with the entire dataset comprising over 10.5 million text samples and more than 5.5 billion words.}
 }
	\label{fig:cloud}
% \vspace{-15pt}
\end{figure}

\begin{itemize}
\item [1)] 
\textbf{Geographical Coverage}:
We performed a statistical analysis of the geographical distribution of images in the Git-10M dataset. As shown in Fig. \ref{fig:Geo_distribution}, Git-10M spans multiple continents and geographical regions, covering various typical scenes such as urban areas, forests, mountains, deserts, and more. The wide geographical coverage ensures that the dataset can support the generation of real-world remote sensing images across different regions, natural features, and diverse scenes. Besides, as stated in the Section \ref{Image_Collection}, {some images of our Git-10M dataset are collected from several public scene classification datasets, which provide explicit scene labels. The integration of these datasets ensures that Git-10M covers a wide variety of well-defined remote sensing scene types. For example, the AID dataset contains 30 typical scene categories, while the RSICB dataset contains 45 categories.}

\item [2)] 
\textbf{Resolution Distribution}:
Fig. \ref{fig:resolution} illustrates the distribution of images with varying resolutions in the Git-10M dataset. The dataset encompasses images ranging from high resolution (e.g., 0.5m/pixel) to low resolution (e.g., 128m/pixel). High-resolution images capture detailed features, making them suitable for tasks that require fine-grained information. On the other hand, low-resolution images provide a broader coverage of larger areas. The multi-resolution nature of the Git-10M dataset offers essential support for training models that can generate images at specific scales.

% thereby enhancing the model's ability to handle various spatial contexts.

\begin{figure}
	\centering
 % \vspace{-15pt}
	\includegraphics[width=1\linewidth]{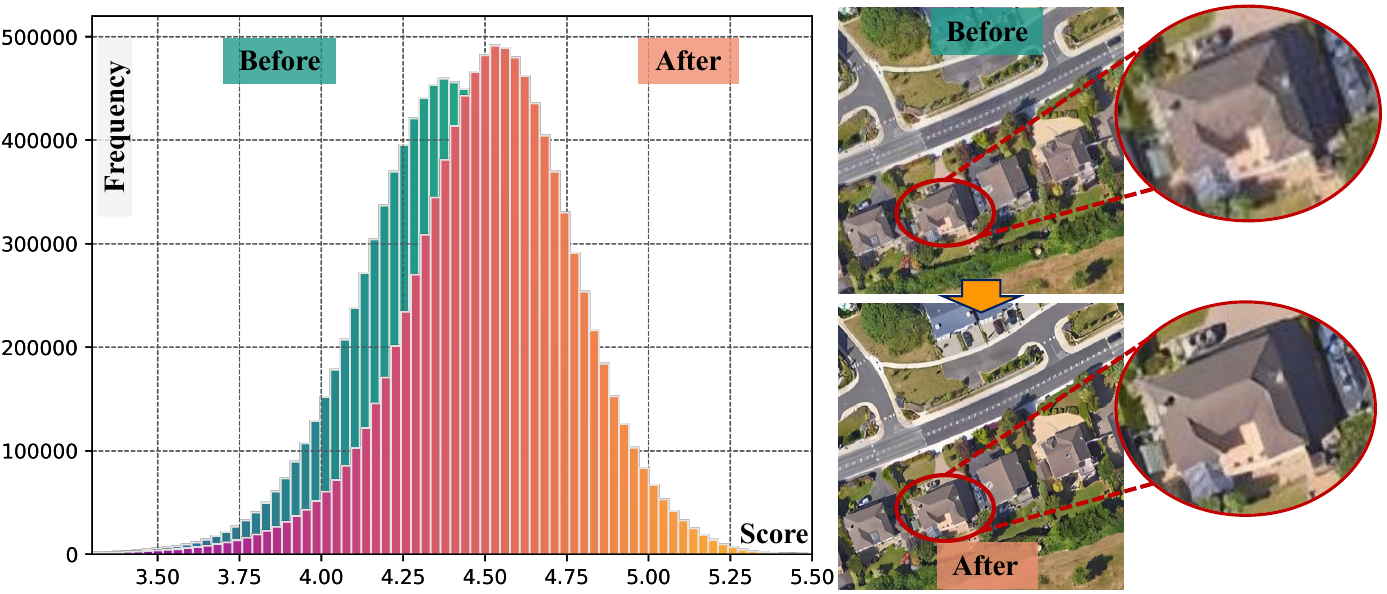}
        % \vspace{-10pt}
	\caption{{The quality score of images before and after enhancement processing for our Git-10M dataset. The results demonstrate a significant improvement after enhancement. An example is shown on the right.}
 }
	\label{fig:score}
% \vspace{-15pt}
\end{figure}

\item [3)] 
\textbf{Image Evaluation}:
To assess the effectiveness of our image enhancement model, we employed a widely used aesthetic model\footnote{https://github.com/christophschuhmann/improved-aesthetic-predictor} to evaluate the quality of images before and after image enhancement processing. As shown in Fig. \ref{fig:score}, the results demonstrate a significant image quality improvement after enhancement. High-quality images enhance the dataset's visual appeal and provide reliable training data for the generative models. 
% By improving the quality of the collected images, we ensure that the dataset can support high-quality image generation.
% tasks, resulting in more accurate and realistic outputs.

% a widely used aesthetic model to perform quality assessment on images before and after enhancement.

\item [4)] 
\textbf{Text Analysis}:
We conducted a word cloud analysis on the texts in the Git-10M dataset, with the results presented in Fig. \ref{fig:cloud}. The word cloud highlights the richness and diversity of the textual descriptions, indicating the comprehensive range of concepts and objects covered. Besides, we also examined the distribution of text lengths (see Fig. \ref{fig:cloud}). Each image is associated with a text of approximately 52 words on average, totaling more than 10.5 million text samples and over 5.5 billion words across the entire dataset. 
% This demonstrates the richness of the text descriptions, which not only accurately reflect the content of the images but also provide valuable contextual information to support the generation of remote sensing images based on text prompts.

\end{itemize}

In summary, the Git-10M dataset exhibits significant advantages in terms of geographical diversity, resolution distribution, image quality, and the richness of textual descriptions. These characteristics make it an invaluable resource for advancing remote sensing image generation research.

\section{Text2Earth Foundation Model}
Building on the proposed Git-10M dataset, we developed Text2Earth, a 1.3 billion parameter generative foundation model tailored for large-scale remote sensing text2image generation. This section details the model structure and a dynamic condition adaptation strategy for training and inference.

\subsection{Structure of Text2Earth Model}
\begin{figure}
	\centering
 % \vspace{-15pt}
	\includegraphics[width=\linewidth]{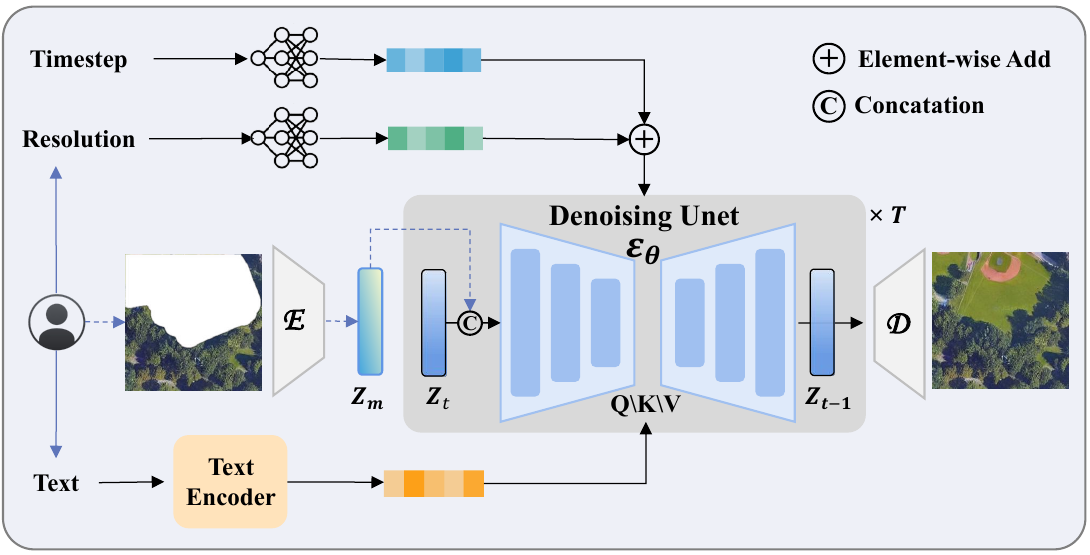}
        % \vspace{-10pt}
	\caption{The structure of the Text2Earth model equipped with 1.3 billion parameters. Text2Earth can generate entirely new images consistent with the provided text or perform local editing on existing images while preserving the original structure. Users can input a white mask to specify the image region for generating visual content, which can either encompass the entire image or focus on a specific area.
 }
	\label{fig:Text2Earth}
% \vspace{-15pt}
\end{figure}

The design of an efficient and powerful foundation model is critical to addressing the demands of global-scale remote sensing image generation. Among various generative architectures, diffusion models stand out for their exceptional capability to model complex data distributions. Leveraging this, we propose Text2Earth, a diffusion-based generative foundation model. As illustrated in Fig. \ref{fig:Text2Earth}, the structure of Text2Earth is built upon three core components: image compression encoding, conditional embedding mechanism, and diffusion modelling. The Variational Autoencoder (VAE) is employed for efficient image compression and reconstruction. A U-Net with a cross-attention mechanism is used for multi-step denoising. OpenCLIP ViT-H text encoder~\cite{CLIP} converts text into high-dimensional semantic embeddings. A resolution embedding module aims to encode image resolution as an implicit embedding. The text embeddings and resolution embedding will be incorporated into each denoising step of the diffusion process.

Our Text2Earth can generate entirely new remote sensing images consistent with the provided text and resolution or perform local editing on existing images while preserving the original structure. Users can input a white mask to specify the image region for generating visual content, which can either encompass the entire image or focus on a specific area.

% Image implicit Encoding: Efficient compression and reconstruction of remote sensing images via a Variational Autoencoder (VAE).
% Conditional Embedding Mechanism: High-dimensional semantic embeddings derived from textual input using an OpenCLIP ViT-H encoder~\cite{CLIP}.
% implicit Diffusion Modeling: A U-Net architecture with cross-attention mechanisms for precise noise prediction and generation control.

\subsubsection{\textbf{Image Compression Encoding}}
The VAE is employed to compress high-resolution remote sensing image pixels into a compact implicit space while preserving perceptual consistency between the implicit and pixel spaces~\cite{StableDiffusion}. This significantly enhances computational efficiency for the subsequent diffusion modelling, which is crucial for unbounded and large-scale remote sensing image generation.

Given an input image \( x \in \mathbb{R}^{H \times W \times C} \), the encoder \( \mathcal{E} \) compresses it into an implicit representation \( z \in \mathbb{R}^{h \times w \times c} \), where \( h, w \textless H, W \), thus reducing the dimensionality of the implicit space compared to the original image pixel space. The compression encoder involves multi-scale feature extraction with progressive downsampling, ensuring a compact yet information-rich implicit representation.
% employs multi-scale feature extraction and multiple downsampling factors to control the compression rate of the implicit space.
% using multiple downsampling factors \( f = 2^m \, (m \in \mathbb{N}) \).
The decoder \( \mathcal{D} \) subsequently reconstructs the image \( \hat{x} \) from the implicit representation \( z \) as follows:
\[
\hat{x} = \mathcal{D}(z),\quad \text{where} \quad \hat{x} \approx x.
\]
% The training objective is to minimize the reconstruction loss while ensuring visual quality through perceptual loss.
% The VAE is trained to minimize reconstruction error while maintaining high visual fidelity, facilitated by a perceptual loss function.

\subsubsection{\textbf{Diffusion Modeling}}
The diffusion modeling is at the heart of Text2Earth, enabling high-quality and diverse image generation. 
% Unlike conventional diffusion models operating in pixel space, Text2Earth performs the diffusion process in the compressed implicit space, significantly reducing computational complexity.
% By performing the diffusion modeling in this feature space instead of conventional pixel space, Text2Earth significantly reduces computational overhead while preserving the image fidelity, making it well-suited for unbounded large-scale scene generation
The forward diffusion process gradually corrupts the implicit representation \( z_0 \) by adding Gaussian noise at the timestep \( t \):
\[
z_t = \sqrt{\bar{\alpha}_t} z_0 + \sqrt{1 - \bar{\alpha}_t} \epsilon,
\]
where \( \epsilon \sim \mathcal{N}(0, 1) \) represents Gaussian noise, and \( \bar{\alpha}_t \) is the cumulative scaling factor, which is defined as the product of individual scaling factors \( \alpha_i \) up to timestep \( t \). Mathematically, it is defined as:
\[
\bar{\alpha}_t = \prod_{i=1}^t \alpha_i.
\]

The reverse diffusion process aims to denoise the implicit representation \(z_t\) and reconstruct \(z_0\). The U-Net denoising network \( \epsilon_\theta \) is trained to predict the noise component \( \epsilon \) using the following loss function:
\[
\underset{\theta}{\min} \mathcal{L}_{\text{LDM}} = \mathbb{E}_{z_0, \epsilon \sim \mathcal{N}(0,1), t} \left[ \| \epsilon - \epsilon_\theta(z_t, t, \tau, \rho ) \|^2 \right],
\]
where \( \tau \) represents the semantic embedding derived from text, and \( \rho \) denotes the resolution embedding.
% \( \epsilon_\theta \) is the denoising U-Net network, which takes the noisy implicit representation \( z_t \), timestep \( t \), the semantic embedding \( \tau \) from text, and the resolution embedding \( \rho \) as input.
The well-trained diffusion models generate samples by progressive denoising from Gaussian noise.

By performing diffusion modeling in the VAE's compressed feature space, Text2Earth achieves a substantial reduction in computational requirements while preserving image fidelity. This makes Text2Earth suitable for large-scale and unbounded remote sensing image generation.

\subsubsection{\textbf{Conditional Embedding Mechanism}}
The conditional embedding mechanism in Text2Earth integrates textual semantics and resolution control at each step of the reverse diffusion process. This will guide noise prediction and ensures that the generated image aligns with both the textual description and the specified resolution, achieving precise and customizable image generation.

% \textbf{Text Embedding}:
Text2Earth utilizes the OpenCLIP ViT-H text encoder \( \mathcal{T} \)~\cite{CLIP} to transform the input text \( I_t \) to a high-dimensional semantic embedding \( \tau \):
\[
\tau = \mathcal{T}(I_t), \quad \tau \in \mathbb{R}^{L \times d}.
\]
where \(L\) is the token length and of \( d \) is the embedding dimension. To effectively incorporate semantic information to guide visual content generation, Text2Earth employs a cross-attention mechanism, injecting the text embedding \( \tau \) into the intermediate layers of the denoising U-Net. The cross-attention mechanism is defined as:
\[
\text{Attention}(Q, K, V) = \text{Softmax}\left(\frac{QK^\top}{\sqrt{d_k}}\right)V,
\]
where \( Q \) is derived from the noisy implicit representation \( z_t \), and \( K \) and \( V \) come from the text embedding \( \tau \). The scaling factor \( d_k \)  ensures numerical stability. This mechanism enables the model to dynamically focus on critical semantic features in the text, ensuring the generated image is semantically faithful to the textual description.

% \textbf{Resolution Embedding}:
To address the limitations of previous models in resolution control, Text2Earth introduces a resolution guidance mechanism that allows for flexible control over image resolution.
Specifically, resolution information \( I_s \) is encoded into the implicit space using a projection layer, producing a resolution embedding \( \rho \). \( \rho \) is then combined with the timestep embedding \( g_\theta(t) \) as follows:
% and passed to the denoising U-Net network, which adjusts the resolution of the generated image at each diffusion step. 
\[
c_{st} = \rho +g_\theta(t) = f_\theta(I_s)+g_\theta(t)
\]
This embedding is then input to the U-Net, which adjusts the generated image resolution at each diffusion step.
% ensuring that the generated image adheres to the specified resolution while maintaining structural and visual coherence.

% \subsubsection{Masked Image Encoding}
Furthermore, to extend the capabilities of Text2Earth to text-driven image editing tasks, a conditional masked image encoding mechanism is introduced. 
Specifically, given an input-masked image \( x_m \in \mathbb{R}^{H \times W \times C} \), the VAE encoder \( \mathcal{E} \) generates the implicit representation \( z_m \in \mathbb{R}^{h \times w \times c_m} \). The \( z_m \) is then concatenated with the implicit variable \( z_t \in \mathbb{R}^{h \times w \times c} \) obtained from the diffusion process along the channel dimension to form a joint conditional representation:
\[
z_{\text{cond}} = [z_m, z_t] \in \mathbb{R}^{h \times w \times (c+c_m)}.
\]
The \( z_{\text{cond}} \) is then passed into the denoising U-Net for noise prediction. This mechanism enables Text2Earth to not only generate entirely new remote sensing images consistent with the provided text and resolution but also perform local editing on existing images while preserving the original structure. 
For example, when certain regions of an image are masked, the model can generate coherent and natural restorations or modifications consistent with the input textual instructions. This capability broadens its applicability to scenarios requiring fine-grained image editing.

\subsection{Dynamic Condition Adaptation Strategy}
\label{section:Dynamic Condition Adaptation Approach}
% To improve the robustness of the Text2Earth model, we propose a dynamic condition adaptation approach for both training and inference. It is inspired by the Classifier-Free Guidance (CFG) strategy~\cite{ho2022_CFG}. 
% During training, text and resolution conditions are randomly dropped with a predefined probability. This enables the model to function effectively even when text or resolution inputs are absent. In the inference phase, the strategy combines the conditional input with a null condition to guide the denoising direction. This allows the model to generate images that closely align with the desired input conditions while maintaining diversity and quality.
% enabling the model to generate images effectively even when input information is partially missing. This method improves the model's flexibility and performance in real-world scenarios. 

To enhance the robustness and adaptability of the Text2Earth model, we propose a Dynamic Condition Adaptation (DCA) strategy. This strategy enables consistent and high-quality image generation. Besides, it can improve the model's adaptability when conditional inputs, such as text or resolution, are missing. The DCA approach involves two key phases: training with dynamic conditioning and sampling with scalable condition guidance.

\begin{algorithm}[t]
  \caption{Training with Dynamic Conditioning} \label{alg:training_dcca}
  \small
  \begin{algorithmic}[1]
    \Repeat
      \State $\bx_0 \sim q(\bx_0)$ \quad \text{(Sample an image from the data distribution)}
       \State $\bz_0 = \mathcal{E}(\bx_0)$ \quad \text{(VAE encoding)}
      \State $t \sim \mathrm{Uniform}(\{1, \dotsc, T\})$ \quad \text{(Random time step)}
      \State $\bepsilon \sim \mathcal{N}(\bzero, \bI)$ \quad \text{(Sample Gaussian noise)}
      \State $c_{\text{text}} \sim \text{Bernoulli}(p_{1})$ \quad \text{(Randomly drop text: 0 or 1)}
      \State $c_{\text{res}} \sim \text{Bernoulli}(p_{2})$ \quad \text{(Randomly drop resolution: 0 or 1)}
      \State $I_t \leftarrow $ \text{the corresponding text of} $\bx_0$
      \State $I_s \leftarrow $ \text{the corresponding resolution of} $\bx_0$
      \If{$c_{\text{text}}$ ==1}
        \State $\tau \leftarrow \tau_\varnothing$ \quad \text{(unknown text embedding)}
      \Else
        \State $\tau \leftarrow \mathcal{T}(I_t)$ \quad \text{(text embedding)}
      \EndIf
      \If{$c_{\text{res}}$ ==0}
        \State $\rho \leftarrow \rho_\varnothing$ \quad \text{(unknown resolution embedding)}
      \Else
        \State $\rho \leftarrow f_\theta(I_s)$ \quad \text{(resolution embedding)}
      \EndIf
      \State Take gradient descent step on
      \Statex $\qquad \grad_\theta \left\| \bepsilon - \bepsilon_\theta(\sqrt{\bar\alpha_t} \bz_0 + \sqrt{1-\bar\alpha_t} \bepsilon, t, \tau, \rho) \right\|^2$
    \Until{converged}
  \end{algorithmic}
\end{algorithm}

\subsubsection{\textbf{Training with Dynamic Conditioning}}
During training, text and resolution conditions are randomly dropped with predefined probabilities. This strategy encourages the model to learn denoising dynamics and feature representations that are robust to incomplete or missing conditions, simulating real-world scenarios where inputs might be absent or unreliable. The training procedure incorporates both conditional and unconditional learning. When text and resolution conditions are present, the model learns to generate images that align closely with these inputs. When both conditions are dropped, the model learns to generate images based purely on noise, akin to traditional unconditional generation.
This dynamic conditioning process ensures that Text2Earth can handle a wide range of input scenarios, enhancing its flexibility and robustness. The training steps are detailed in Algorithm \ref{alg:training_dcca}.
% as shown in Algorithm \ref{alg:training_dcca}. The training process involves randomly applying condition dropout during each training iteration. The model learns to generate both conditional and unconditional image generation. In the absence of both text and resolution conditions, the model learns to generate images based purely on noise, akin to traditional unconditional image generation models.

\subsubsection{\textbf{Sampling with Scalable Condition Guidance}}
\label{section: Sampling with Guidance}
% In the inference phase, the strategy combines the conditional input with a null condition to guide the denoising direction. This allows the model to generate images that closely align with the desired input conditions while maintaining diversity and quality.

During sampling, the DCA strategy leverages a mixture of conditional input and a null condition to refine the image generation process. This combination guides the denoising process to align generated images closely with the desired conditions while maintaining diversity and quality. Inspired by the classifier-free guidance technique~\cite{ho2022_CFG}, the Text2Earth model predicts two versions of the noise at each denoising step: one conditioned on the input and one without conditioning. The final predicted noise \( \epsilon_{\text{g}} \) is computed as a weighted combination of these two predictions:
% obtained by combining these two predictions, with a weight factor that controls how strongly the model adheres to the provided conditions. This process is formally expressed as follows:
\[
\epsilon_{\text{g}} = (1 + \omega) \epsilon_\theta(z_t, t, \tau, \rho) - \omega \epsilon_\theta(z_t, t, \tau_\varnothing, \rho_\varnothing)
\]
where \(\omega\) is a guidance scale factor that controls the model's reliance on the provided conditions. The sampling process is formalized in Algorithm \ref{alg:sampling_dcca}.

\begin{algorithm}[t]
  \caption{Sampling with Scalable Condition Guidance} \label{alg:sampling_dcca}
  \small
  \begin{algorithmic}[1]
    \State $\bx_T \sim \mathcal{N}(\bzero, \bI)$ \quad \text{(Start with Gaussian noise)}
    \State $I_t \leftarrow $ \text{the input text} 
    \State $I_s \leftarrow $ \text{the input resolution}
    \State $\tau \leftarrow \mathcal{T}(I_t)$ \quad \text{(text embedding)}
    \State $\rho \leftarrow f_\theta(I_s)$ \quad \text{(resolution embedding)}
    \For{$t=T, \dotsc, 1$}
      \State $\bz \sim \mathcal{N}(\bzero, \bI)$ if $t > 1$, else $\bz = \bzero$
       \State $\epsilon_{\text{g}} = (1 + \omega) \epsilon_\theta(z_t, t, \tau, \rho) - \omega \epsilon_\theta(z_t, t, \tau_\varnothing, \rho_\varnothing)$
      \State $\bz_{t-1} = \frac{1}{\sqrt{\alpha_t}} \left( \bz_t - \frac{1-\alpha_t}{\sqrt{1-\bar\alpha_t}} \bepsilon_\theta(\bz_t, t, \tau) \right) + \sigma_t \bz$
    \EndFor
    \State $\bx_0 \leftarrow \mathcal{D}(\bz_{0})$ \quad \text{(VAE decoding)}
    \State \textbf{return} $\bx_0$ \quad \text{(Final generated image)}
  \end{algorithmic}
\end{algorithm}

In summary, the DCA strategy equips Text2Earth with the ability to handle various input scenarios effectively, such as incomplete input conditions. Besides, this strategy facilitates the model to generate images that closely align with the input conditions while maintaining diversity and quality.

\section{Experiment}
% This section

\subsection{Dataset}
\subsubsection{Git-10M Dataset}
The Git-10M dataset comprises 10 million global remote sensing image-text pairs, spanning diverse geographical locations and environmental conditions. This extensive dataset offers a robust foundation for training models capable of generating high-quality, diverse remote sensing imagery.

\subsubsection{RSICD Dataset}
RSICD dataset is a widely used benchmark dataset for remote sensing text2image generation. It contains 10,921 remote sensing images and corresponding text annotations. The dataset contains 30 types of common ground scenes, and the spatial resolution of images is not unique. This dataset was employed to evaluate our model's adaptation to the small specific scene dataset. To further explore the multimodal image generation, we extend the RSICD dataset to a multimodal dataset. RGB images in the RSICD dataset were transformed into various modalities as follows:
\begin{itemize}
\item
Panchromatic (PAN) Images: Converted from original RGB images using grayscale transformation to simulate monochromatic imagery.
\item
Near-Infrared (NIR) Images: Generated using pretrained models to simulate spectral information beyond the visible spectrum.
\item
Synthetic Aperture Radar (SAR) Images: Produced using a pretrained model based on the Pix2Pix framework, providing radar-like image representations.
\item
Low-Resolution Images: Obtained by downsampling RGB images, simulating scenarios with constrained spatial resolutions.
\item
Foggy Images: Synthesized by adding fog to the original image using a classic fog simulation algorithm.
\end{itemize}

\subsection{Implementation Details}
Distributed training was conducted on a machine equipped with 8 NVIDIA A100 GPUs to manage the computational demands of training large-scale generative models. The training setup utilized the AdamW optimizer with a learning rate of \(0.0001\), and a batch size of \(1024\) was chosen to maximize hardware utilization and ensure efficient gradient updates. The generated image size is set to \(256 \times 256\) pixels.

A progressive training strategy was used to improve the model's ability to generate diverse and high-quality remote sensing images. The model was initially trained on the complete Git-10M dataset, leveraging its extensive diversity to capture a wide range of spatial and spectral geographic features. The model was subsequently fine-tuned on a high-quality subset of the dataset, comprising samples with a score greater than 4.8 in Fig. \ref{fig:score}. This refinement phase improved the fidelity and detail of the generated images. This two-stage approach allowed the model to learn from a broad dataset and refine its generation capabilities on a higher-quality sub-set.

We developed two specialized versions of the Text2Earth model to address distinct remote sensing tasks. Text2Earth$_{t}$ was optimized for generating remote sensing images from text and resolution. Text2Earth$_{e}$ was tailored for image editing tasks. This flexibility allows Text2Earth to cater to a wide range of practical remote sensing applications.
% from generating fully synthetic data to repairing and editing existing imagery.

\subsection{Evaluation Metrics}
The Fréchet Inception Distance (FID) metric is widely used to evaluate generative models by measuring the perceptual similarity between generated and real images. It compares the distributions of features extracted from both sets in a shared feature space. A lower FID score indicates better quality and diversity of the generated images. 
The FID score is computed as follows:
\[
\text{FID} = \|\mu_r - \mu_g\|_2^2 + \text{Tr}(\Sigma_r + \Sigma_g - 2{(\Sigma_r \Sigma_g)}^{1/2})
\]
where \( \mu_r \) and \( \Sigma_r \) denote the mean and covariance of features extracted from the real image distribution.
\( \mu_g \) and \( \Sigma_g \) are the mean and covariance of features extracted from the generated image distribution, respectively. \(\text{Tr}\) denotes the trace of a matrix.
The features for FID calculation are extracted from a pre-trained Inception-v3 network~\cite{szegedy2016rethinking_inception}, ensuring a perceptually relevant image representation.

Following previous studies~\cite{Txt2Img_MHN,txt2rs3_CRS_diff}, we also employ the Zero-Shot classification Overall Accuracy (Cls-OA) metric to evaluate the semantic alignment between generated images and their textual descriptions. Specifically, {a classification model (i.e., ResNet-18) is trained on generated images using text descriptions from the test set. This model is then used for zero-shot classification on the real test set without prior exposure to them during training. The OA metric thus measures the semantic coherence and relevance of the generated images to the textual prompts.}

{Furthermore, to measure the semantic similarity between text and generated images, we also used the CLIP score, which is calculated as follows:}
\[
\text{CLIP Score} = \frac{1}{N} \sum_{i=1}^{N} \cos \left( E_\text{image}(I_i), E_\text{text}(T_i) \right) \times 100
\]
where \(I_i\) denotes the \(i\)-th generated image, \(T_i\) denotes the corresponding input text, \(E_\text{image}\) and \(E_\text{text}\) represent the image encoder and text encoder of a pretrained CLIP model, respectively, and \(\cos(\cdot)\) denotes the cosine similarity between the two embedding vectors. The final CLIP score is the average cosine similarity across all \(N\) text-image pairs, reflecting the overall semantic alignment quality.

% This metric measures the mean cosine similarity between the text descriptions and the generated images in the feature space of a CLIP model. The CLIP-RSICD-V2model, fine-tuned on the RSICD dataset, is adopted in the evaluation.

% , providing a robust indicator of model performance.

% we also adopt Zero-shot overall accuracy (OA) metric. emphasizes the alignment between the generated images and their textual descriptions. Specifically, a classification model (e.g., ResNet-18) is trained on the generated images using text descriptions from the test set. This model is then used for zero-shot classification on the real test set, leveraging the category labels inherent to datasets such as RSICD. Since the classification model is not exposed to real test images during training, the OA metric effectively measures the semantic coherence and relevance of the generated content to the textual prompts.

\begin{table}%[htbp]%[!t]  %\small
\renewcommand{\arraystretch}{1.3}
\caption{{Comparisons between our Text2Earth model and previous text2image methods on the RSICD dataset.}}
\label{tab:method_RSICD_cpmparison}
\centering
% \resizebox{0.9\linewidth}{32mm}{
% \begin{tabular}{c |c c c c}
\centering
\begin{tabular}{m{85pt}<{\centering}|m{30pt}<{\centering}m{40pt}<{\centering}m{45pt}<{\centering}}
	\toprule%[1pt]
        Method & FID $\downarrow$ & Zero-Shot Cls-OA $\uparrow$ & CLIP Score $\uparrow$\\
        \midrule
        Attn-GAN~\cite{xu2018attn_gan} & 95.81 & 32.56\% & 20.19\\
        DAE-GAN~\cite{ruan2021dae_gan} & 93.15 & 29.74\% & 19.69\\
        % StrucGAN~\cite{zhao_StrucGAN}    & –  \\
        DF-GAN~\cite{tao2022df_GAN}        & 109.41 & 51.99\% & 19.76\\
        Lafite~\cite{zhou2022towards_LAFITE}    & 74.11 & 49.37\% & 22.52\\
        % \midrule
        DALL-E~\cite{ramesh2021zero_DALLE}  & 191.93 & 28.59\% & 20.13\\
        Txt2Img-MHN\(_\text{vqvae}\)~\cite{Txt2Img_MHN} & 175.36 & 41.46\% & 21.35\\
        Txt2Img-MHN\(_\text{vqgan}\)~\cite{Txt2Img_MHN}  & 102.44 & 65.72\% & 20.27\\
        % \midrule
        RSDiff~\cite{txt2rs4_rsdiff} & 66.49 & -- & --\\
        CRS-Diff~\cite{txt2rs3_CRS_diff} & {50.72} & 69.31\% & 20.33\\
        \midrule
        Text2Earth (Ours) & \textbf{24.49} & \textbf{90.26\%} & \textbf{25.62}\\
	\bottomrule
\end{tabular}
% }
\end{table}

\begin{figure*}
	\centering
 % \vspace{-20pt}
	\includegraphics[width=1\linewidth]{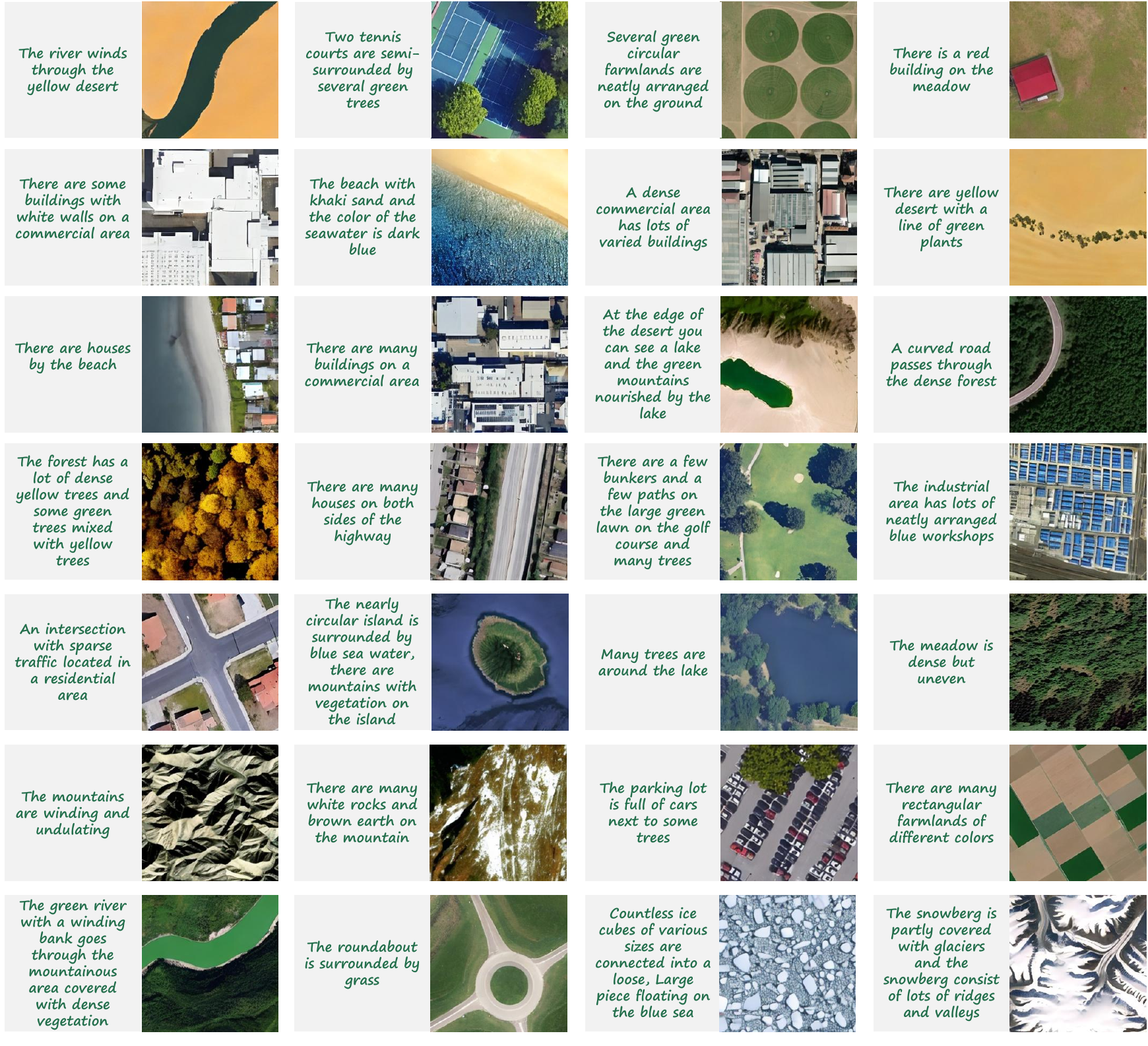}
    \includegraphics[width=1\linewidth]{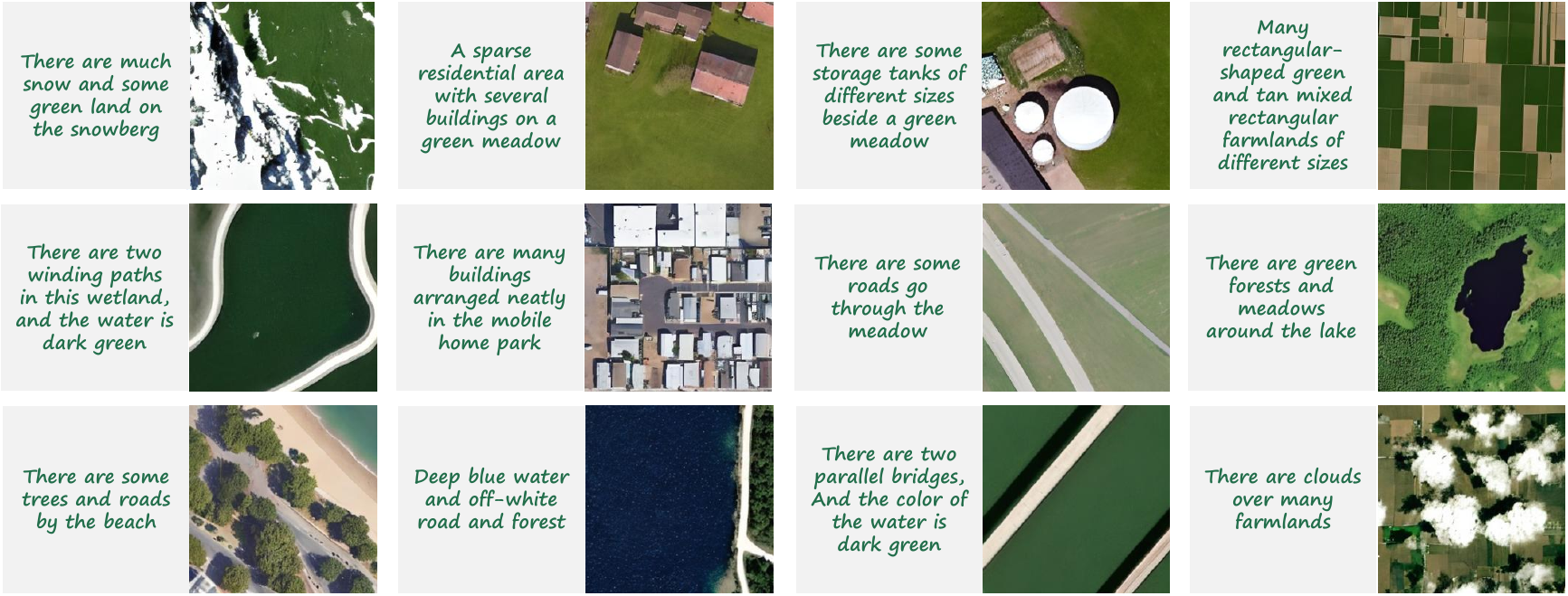}
        % \vspace{-15pt}
	\caption{Our Text2Earth demonstrates robust capabilities for zero-shot text2image generation across diverse geographical features based on user-free text input. It can generate a variety of scenes, including diverse geographical features such as mountain ranges, rivers, urban areas, forests, and farmland.  
 }
	\label{fig:Text2Img_result}
% \vspace{-15pt}
\end{figure*}

\begin{figure*}
	\centering
 % \vspace{-20pt}
	\includegraphics[width=1\linewidth]{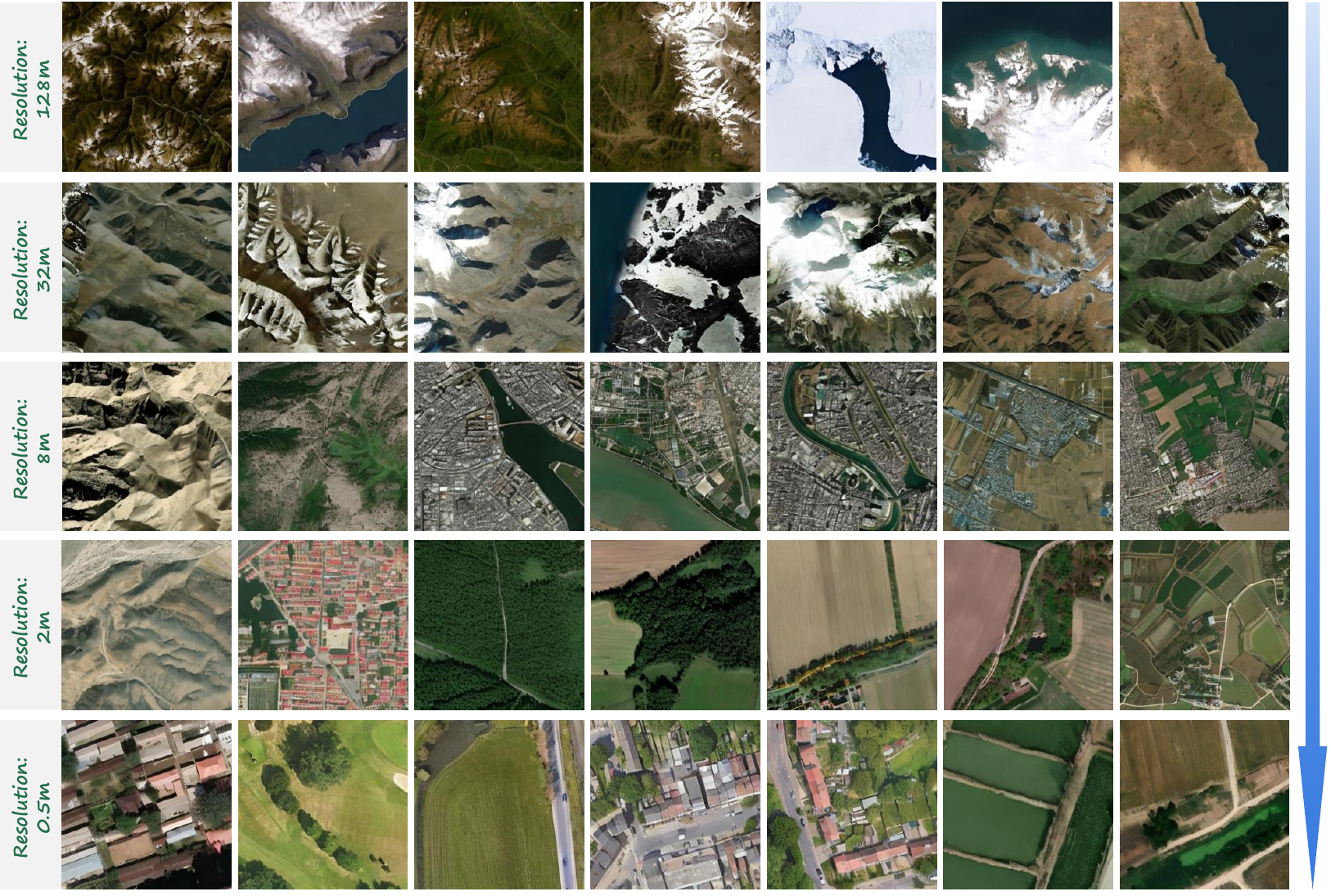}
        % \vspace{-15pt}
	\caption{{Generated images with different resolutions solely by specifying the resolution condition, without any descriptive text.. Text2Earth can generate images reflecting a range of spatial resolutions—from high-resolution close-up views that capture fine details to lower-resolution images that cover larger areas. For example, in the generated images of mountainous regions, higher-resolution images exhibit detailed terrain features, while lower-resolution images depict broader landscape coverage, which aligns with real-world spatial resolution characteristics.
    }
 }
	\label{fig:Text2Img_result_res}
% \vspace{-15pt}
\end{figure*}

\begin{figure}
	\centering
 % \vspace{-20pt}
	\includegraphics[width=1\linewidth]{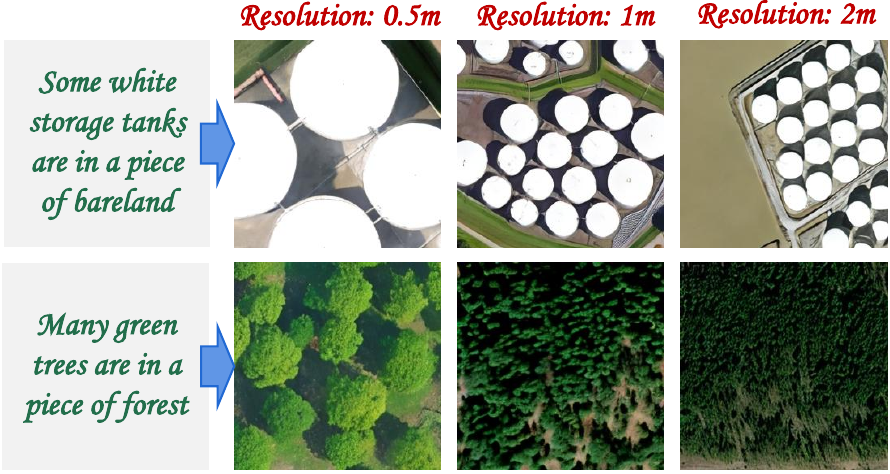}
        % \vspace{-15pt}
	\caption{{Resolution-conditioned image generation with same text prompts.}
 }
	\label{fig:text_dif_res}
% \vspace{-15pt}
\end{figure}

\begin{figure}
	\centering
 % \vspace{-20pt}
	\includegraphics[width=0.8\linewidth]{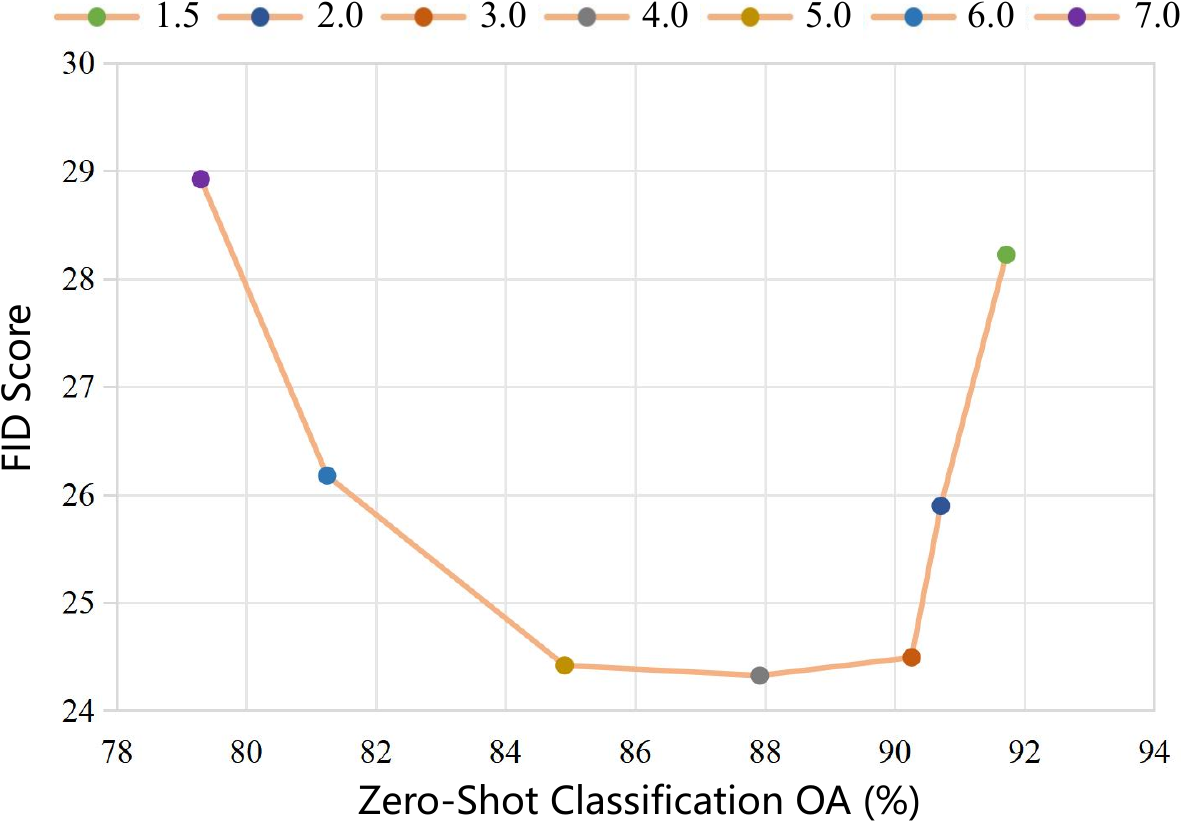}
        % \vspace{-15pt}
	\caption{Evaluation results of our Text2Earth on the RSICD dataset under different guidance scale factors \(\omega\) (i.e., 1.5, 2.0, 3.0, 4.0, 5.0, 6.0, 7.0).
 }
	\label{fig:guidance_scale}
% \vspace{-15pt}
\end{figure}

\subsection{Zero-Shot Text2Image Generation}
Different from previous methods that are limited to generating images for specific scenes, Text2Earth is trained on our large-scale dataset, endowing it with robust capabilities for zero-shot text2image generation across a wide range of geographical and environmental features. It can generate specific image content based on user-free text input, without scene-specific fine-tuning or retraining. As shown in Fig. \ref{fig:Text2Img_result}, Text2Earth can generate a variety of scenes, including diverse geographical features such as mountains, rivers, urban areas, forests, and farmland. Additionally, Text2Earth is capable of generating remote sensing images at various resolutions based on user specifications. For example, as shown in Fig. \ref{fig:Text2Img_result_res}, it can generate high-resolution images of urban landscapes with detailed buildings or low-resolution images depicting expansive forest covers. This versatility demonstrates the model's ability to adapt to varying input conditions and user requirements.

% \subsubsection{Handling Missing Conditions: Robust Generation}
Text2Earth also demonstrates remarkable robustness in generating realistic images, even in cases where key input conditions—such as text or resolution—are missing. For instance, the model can generate forest images at various scales when provided with text like ``There is a dense forest" without a specified resolution. Besides, when only the resolution is provided without specific textual descriptions, the model can generate resolution-specific images with diverse scenes such as a forest or an urban area. This ability highlights the model’s robustness in handling incomplete or missing input data, making it adaptable for real-world applications where input conditions may be partial. This ability benefits from our proposed dynamic condition adaptation strategy described in Section \ref{section:Dynamic Condition Adaptation Approach}.

In Fig. \ref{fig:text_dif_res}, {we tested whether the model could generate images with different levels of spatial detail given the same textual description but different resolution conditions. For example, using the prompt ``Some white storage tanks are in a piece of bare land," we generated images at 0.5m, 1m, and 2m per pixel resolutions. The resulting images exhibited variations in the relative size and density of the storage tanks that corresponded well with the specified resolutions, effectively mimicking real-world scale variations. Similarly, for the prompt ``Many green trees are in a piece of forest," the level of detail in tree structures varied appropriately with the resolution, demonstrating the model’s capability to produce resolution-dependent images.}

% \subsubsection{Creativity in Image Generation}
% The creativity is an exciting advantage of Text2Earth, setting it apart from previous text2image models. By leveraging a large and diverse training dataset, the model can imaginatively combine remote sensing geographical elements in visually novel ways to generate creative and novel images, which were never explicitly seen during training. This opens up new possibilities for applications in geospatial data generation, simulation, and creative content creation.
% For example, when given the prompt "a house on a road," Text2Earth can generate an image of a house unexpectedly placed on a road. Similarly, when asked for "an oil tank on a baseball field," the model creatively places an industrial oil tank in the middle of a sports field, combining elements that seem improbable but remain visually plausible. These examples demonstrate the model's ability to generate imaginative and novel content, reflecting its flexibility and creativity.

The power of Text2Earth lies in its rich potential knowledge and general image generation capabilities learned from extensive training data, enabling it to adapt to new datasets through fine-tuning quickly. To further validate the robustness of Text2Earth as a foundation model, we fine-tuned it using the Low-Rank Adaptation (LoRA) technique~\cite{hu2021lora} on the widely used remote sensing text2image benchmark dataset RSICD~\cite{Lu_2018_RSICD}. facilitates efficient transfer learning by introducing a small number of learnable low-rank matrices while keeping the original model parameters fixed. As shown in Table \ref{tab:method_RSICD_cpmparison}, Text2Earth significantly outperforms previous methods on the RSICD dataset, achieving a remarkable improvement of +26.23 in FID and +20.95\% in Zero-Shot Classification OA. These improvements demonstrate the robustness of Text2Earth as a foundation model, which can effectively transfer its learned general knowledge to specific tasks through LoRA fine-tuning.

% These results demonstrate the robustness and transferability of Text2Earth as a generative foundation model. confirming its ability to effectively leverage learned general knowledge and adapt to specific tasks through LoRA fine-tuning.

Additionally, we present evaluation results of our Text2Earth on the RSICD dataset under different guidance scale factors \(\omega\) during inference, as shown in Table \ref{tab:guidance_scale} and Fig. \ref{fig:guidance_scale}. When \(\omega\) is set to 3.0, the model achieves a favourable trade-off between FID and Zero-shot Cls-OA, further highlighting its flexibility in balancing image quality and semantic alignment.

\begin{table}%[htbp]%[!t]  %\small
\renewcommand{\arraystretch}{1.3}
\caption{Evaluation results of our Text2Earth on the RSICD dataset under different guidance scale factors.}
\label{tab:guidance_scale}
\centering
% \resizebox{0.9\linewidth}{32mm}{
% \begin{tabular}{c |c c c c}
\centering
\begin{tabular}{m{55pt}<{\centering}|m{50pt}<{\centering}m{100pt}<{\centering}}
	\toprule%[1pt]
        Guidance Scale & FID Score $\downarrow$ & Zero-Shot Classification OA $\uparrow$ \\
        \midrule
        \(\omega\) = 1.5 & 28.22 & 91.72\% \\
        \(\omega\) = 2.0 & 25.89 & 90.71\% \\
        \(\omega\) = 3.0 & 24.49 & 90.26\% \\
        \(\omega\) = 4.0 & 24.32 & 87.92\% \\
        \(\omega\) = 5.0 & 24.42 & 84.91\% \\
        \(\omega\) = 6.0 & 26.17 & 81.25\% \\
        \(\omega\) = 7.0 & 28.92 & 79.30\% \\
	\bottomrule
\end{tabular}
% }
\end{table}

\begin{figure*}
	\centering
 % \vspace{-20pt}
	\includegraphics[width=1\linewidth]{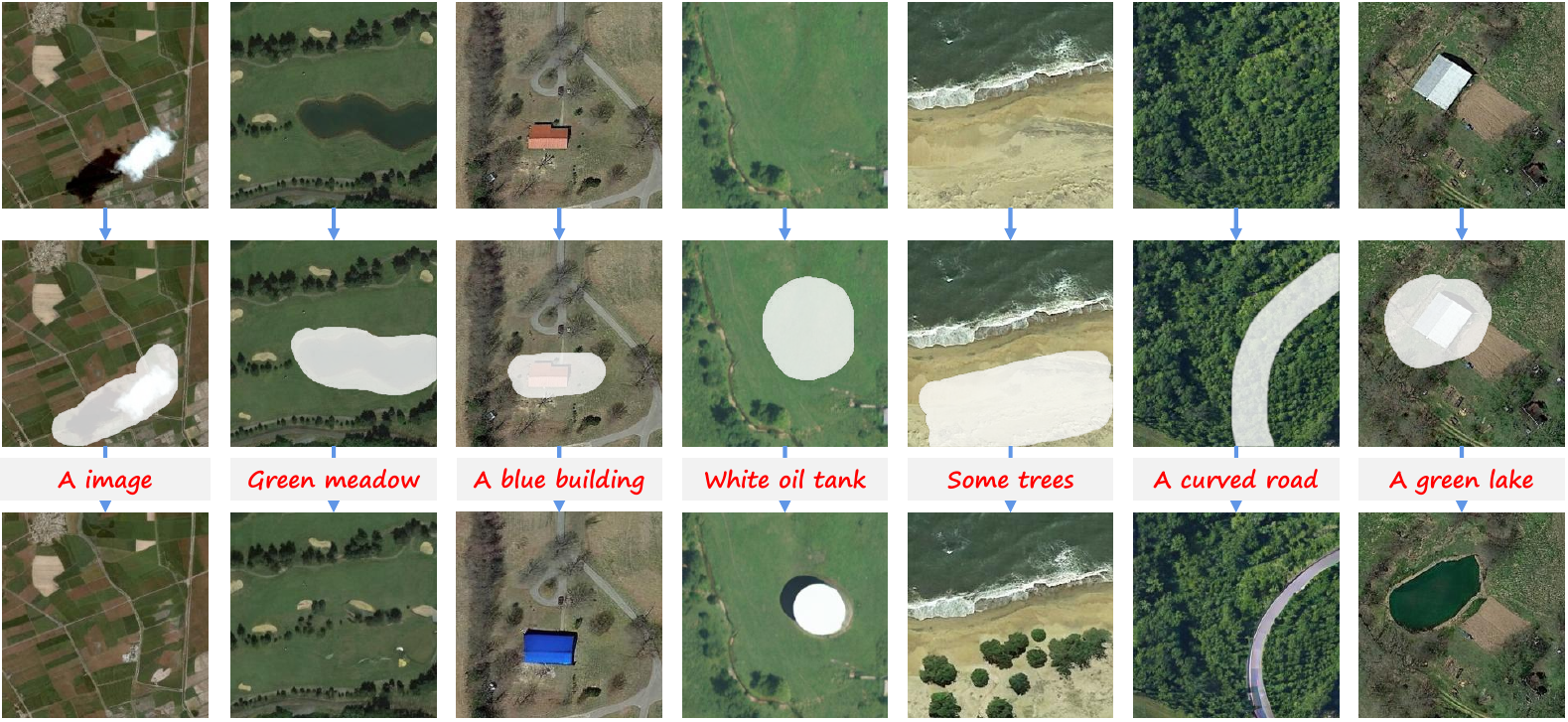}
        % \vspace{-15pt}
	\caption{Some examples in remote sensing image editing. Text2Earth exhibits exceptional versatility in remote sensing image editing, enabling modifications to image content such as removing clouds, and replacing or adding geographic features.
 }
	\label{fig:inpaint}
% \vspace{-15pt}
\end{figure*}

\subsection{Remote Sensing Image Editing}
In addition to text2image generation, Text2Earth exhibits exceptional versatility in remote sensing image editing, enabling modifications to image content such as replacing or removing geographic features. These capabilities are valuable across a range of practical applications, as demonstrated in the examples shown in Fig. \ref{fig:inpaint}. For instance, in the cloud removal example presented in Fig. \ref{fig:inpaint}, Text2Earth is given an input image with cloud-covered regions and a corresponding mask. Text2Earth can understand the semantic structure of the image and successfully reconstruct the cloud-covered areas, ensuring natural scene continuity. This ability to effectively restore occluded regions while maintaining realistic transitions illustrates Text2Earth's strength in image editing tasks.

Moreover, Text2Earth can perform targeted scene modifications based on user-provided text. For example, when given textual prompts alongside region-specific masks, the model can execute complex editing tasks, such as: replacing a lake with grassland, changing the colour of houses from red to blue, placing an oil tank on a meadow, planting trees near the beach, constructing a road through a forest, and replacing a house with a lake.

Importantly, Text2Earth ensures that these modifications are seamlessly integrated with the surrounding areas, maintaining continuity and coherence. This makes it an ideal tool for customized remote sensing image editing, catering to diverse applications such as urban planning.
% and landscape management.

\begin{figure*}
	\centering
 % \vspace{-20pt}
	\includegraphics[width=1\linewidth]{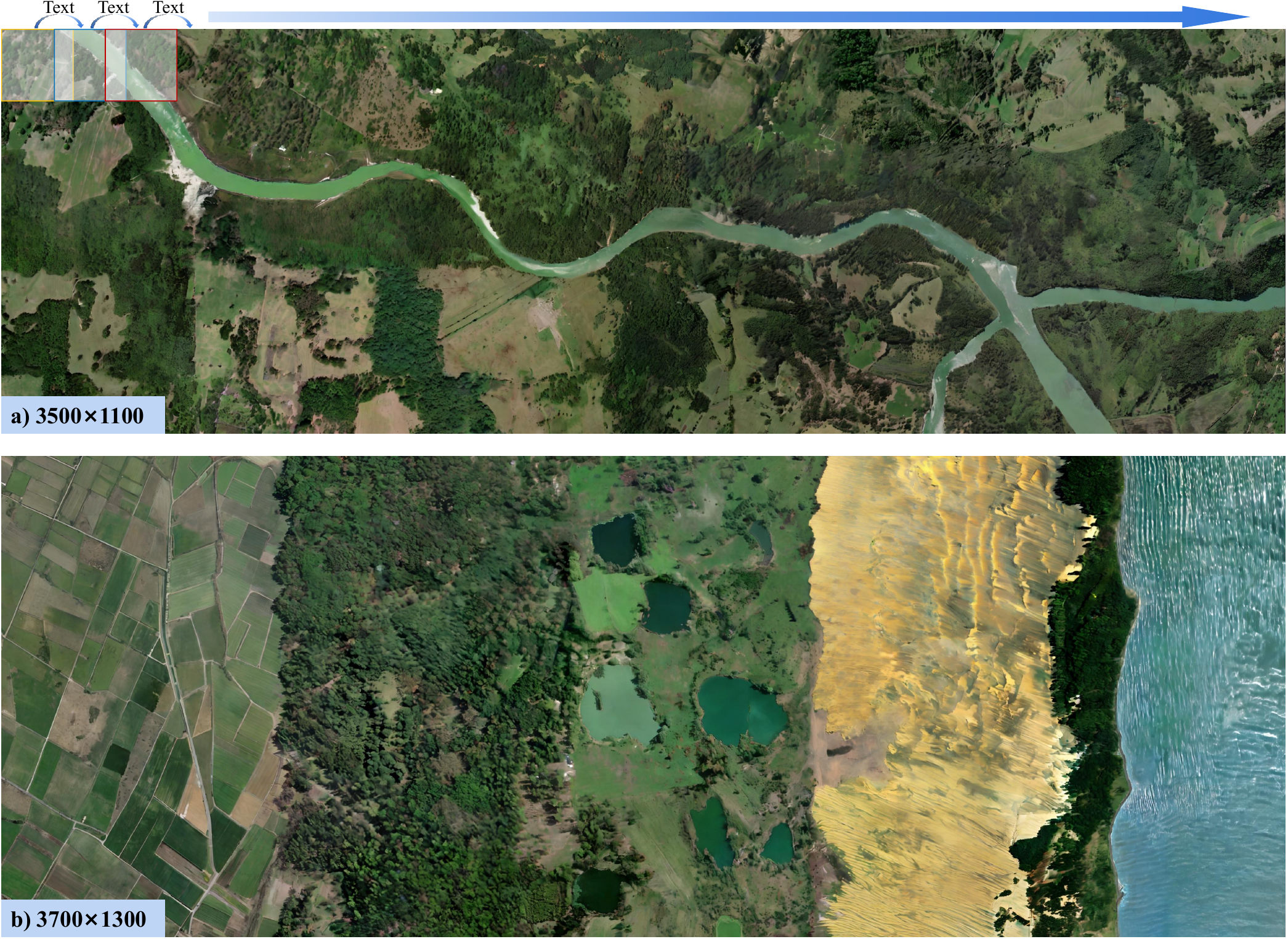}
        % \vspace{-15pt}
	\caption{Unbounded remote sensing scenes through iterative outpainting. Users can seamlessly and infinitely expand remote sensing images on a canvas, effectively overcoming the fixed-size limitations of traditional generative models. 
 }
	\label{fig:unbound}
% \vspace{-15pt}
\end{figure*}

\subsection{Unbounded Remote Sensing Scene Construction}
One of the most innovative applications of Text2Earth is its ability to construct unbounded remote sensing scenes with consistent resolution through iterative outpainting. Using our Text2Earth, users can seamlessly and infinitely generate remote sensing images on a canvas, effectively overcoming the fixed-size limitations of traditional generative models. 
% This technique is particularly valuable for applications requiring the visualization of extensive geographic areas.

The unbounded expansion begins with a base image generated from a user's textual prompt. Users can iteratively provide new textual instructions to guide the content of subsequent image extensions. Text2Earth generates new image segments at the boundaries, ensuring smooth transitions and overall coherence across the expanded scene. We provide two examples in Figure \ref{fig:unbound}. In the first example, we construct a large-scale coverage of the river area, with 3500$\times$1100 pixels, where the river is extended unbounded with some vegetation on both sides. In the second example, we construct a creative large image with seamless transitions between multiple scenes. From the farmland area on the left to the forest, then to the wetland with lakes, then to the desert, followed by some vegetation, and finally transitioning to a blue ocean. 

{Text2Earth’s resolution controllability is the key to maintaining visual coherence across the generated scene during the outpainting process. Using the same resolution at each step, our Text2Earth ensures that different regions of the expanded scene maintain consistent spatial detail. Without such resolution control, varying image resolution across different areas could result in a disjointed or unnatural appearance, undermining the overall coherence of the large scene.}

By generating unbounded scenes with consistent resolution, Text2Earth is valuable for applications requiring the visualization of extensive geographic areas. It will support the creative exploration of spatial planning scenarios, pushing beyond the constraints of traditional workflows.

% By enabling the construction of large-scale scenes, Text2Earth addresses the needs of remote sensing applications requiring expansive geographic visualizations, such as urban development planning and infrastructure design. Its ability to generate unbounded scenes will support the creative exploration of spatial planning scenarios, pushing beyond the constraints of traditional workflows.

\subsection{Cross-Modal Image Generation}
As a powerful generative foundation model, Text2Earth has acquired extensive knowledge and universal image generation capabilities from large-scale remote sensing data. These capabilities not only enable superior performance in text2image generation tasks but also allow for efficient transfer to diverse cross-modal image generation tasks through techniques like parameter-efficient fine-tuning. In this section, we explore Text2Earth’s potential in two key categories of cross-modal image generation tasks.

\begin{figure*}
	\centering
 % \vspace{-20pt}
	\includegraphics[width=1\linewidth]{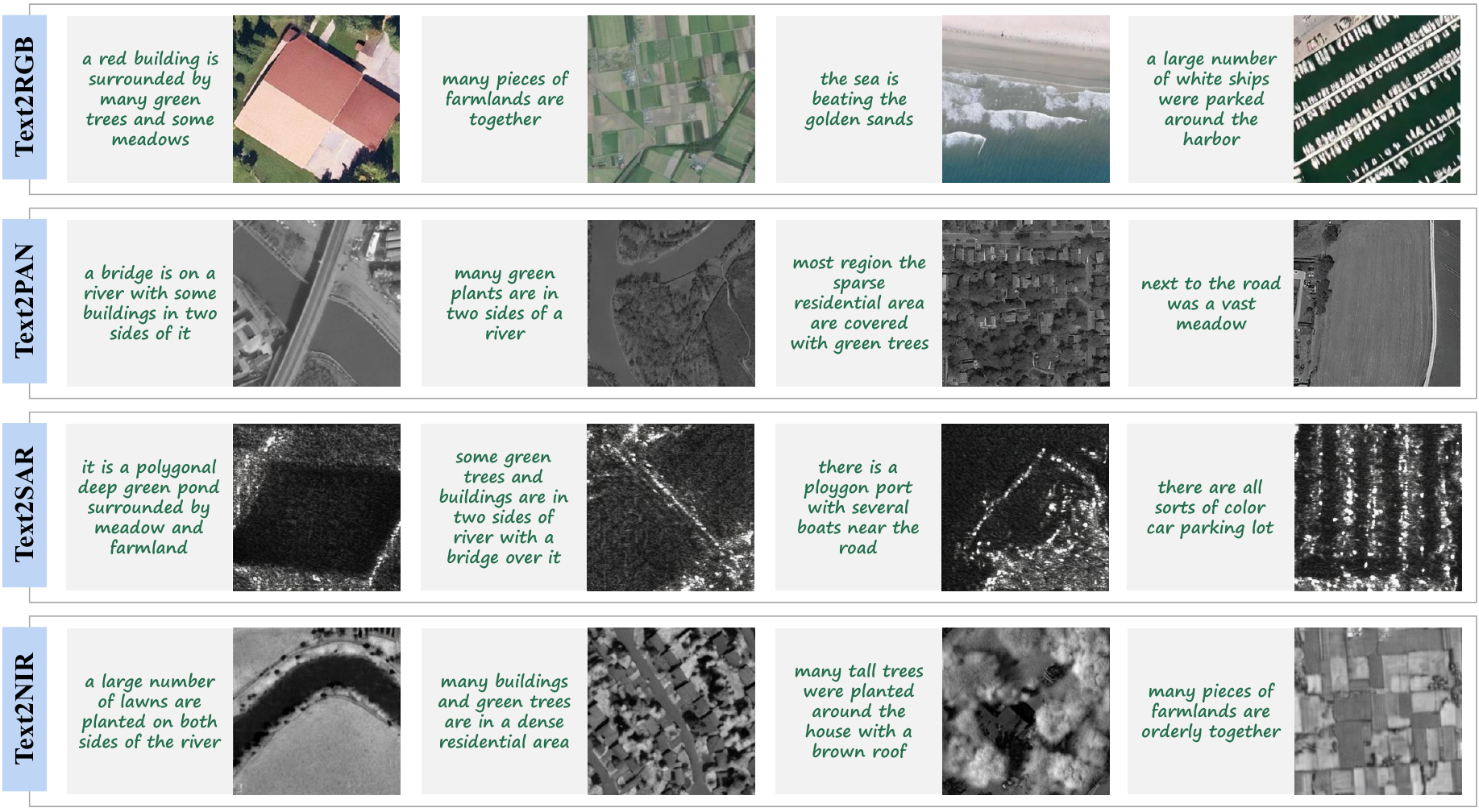}
        % \vspace{-15pt}
	\caption{Text-Driven Multi-Modal Image Generation. Text2Earth can generate high-quality images. For instance, in the generated NIR images, vegetation areas exhibit high pixel values, aligning with the physical imaging principles of NIR, where green vegetation reflects strongly in the near-infrared spectrum.
 }
	\label{fig:text2multimodal}
% \vspace{-15pt}
\end{figure*}

% 在 RSICD 数据集上，我们使用 LoRA 技术对 Text2Earth 进行微调，来实现文本到多模态的图像生成，包括Text2SAR、Text2NIR、Text2PAN。Fig. \ref{fig:Text2Modal} 展示了一些样例，表格5也提供了定量的评价。实验结果表明，Text2Earth 生成的多模态图像在视觉质量和语义一致性上均表现出色，这进一步证明了Text2Earth对多模态数据的迁移能力。例如，对于生成的NIR图像，植被通常表现为较高的像素值，这是符合NIR的物理成像规律的：绿色植被具有高反射率的近红外波段特性。

\subsubsection{\textbf{Text-Driven Multi-Modal Image Generation}}
Text2Earth has gained a profound understanding of image semantics and structural information. It can be used to generate multi-modal remote sensing images, including RGB, SAR, NIR, and PAN images. To achieve this, we employed the LoRA technique~\cite{hu2021lora}, which introduces a small number of learnable low-rank parameters into the model’s attention layers while keeping the pre-trained parameters frozen. This approach offers substantial computational efficiency, making it ideal for resource-constrained environments.

We fine-tuned Text2Earth using LoRA on the RSICD dataset to facilitate text-driven multi-modal image generation tasks, such as Text2SAR, Text2NIR, and Text2PAN. The results are illustrated in Fig. \ref{fig:text2multimodal}. The experiments demonstrate that Text2Earth can generate multi-modal images with high quality and semantic consistency. For instance, in the generated NIR images, green vegetation areas exhibit high pixel values, aligning with the physical imaging principles of NIR, where green vegetation reflects strongly in the near-infrared spectrum. These results underscore Text2Earth's ability to effectively transfer its general knowledge to multi-modal remote sensing image generation tasks.

\begin{table}%[htbp]%[!t]  %\small
\renewcommand{\arraystretch}{1.3}
\caption{Text-Driven Multi-Modal Image Generation. LoRA is used to fine-tune our Text2Earth on the multi-modal image data.}
\label{tab:text2multimodal}
\centering
\centering
\begin{tabular}{m{85pt}<{\centering}|m{50pt}<{\centering}m{75pt}<{\centering}}
	\toprule%[1pt]
        Multi-Modal Generation & FID Score $\downarrow$ & Zero-Shot Cls-OA $\uparrow$ \\
        \midrule
        Text2RGB & 24.49 & 90.26\% \\
        Text2PAN & 4.39 & 88.46\% \\
        Text2SAR & 68.83 & 34.42\% \\
        Text2NIR & 2.05 & 82.08\% \\
	\bottomrule
\end{tabular}
\end{table}

Table \ref{tab:text2multimodal} {shows quantitative evaluation on text-driven multi-modal image generation. FID scores across different modalities are not directly comparable because the FID for each modality is computed using an Inception V3 model pre-trained on data specific to that modality. 
Besides, unlike optical images (RGB, NIR, and PAN), SAR images are captured through microwave radar signals, which makes their visual appearance fundamentally different from optical images. SAR images often contain speckle noise and lack significant color and detailed texture information. These factors reduce the amount of semantic information available for scene classification tasks, rendering scene classification on SAR images inherently more challenging than on RGB, NIR, or PAN images. This leads to a low Zero-Shot Cls-OA score for the Text2SAR generation. In summary, the primary purpose of Table} \ref{tab:text2multimodal} {is to provide baseline benchmark results for future text-driven multi-modal image generation research rather than to directly compare performance across modalities.}

\subsubsection{\textbf{Image-to-Image Translation}}
In addition to text-driven multi-modal generation, Text2Earth also exhibits potential in image-to-image translation tasks, containing cross-modal translation and image enhancement, such as PAN to RGB (PAN2RGB), NIR to RGB (NIR2RGB), PAN to NIR (PAN2NIR), super-resolution, and image dehazing. To implement these tasks, we froze the parameters of the Text2Earth model and incorporated a trainable module inspired by ControlNet~\cite{zhang2023adding_ControlNet} to encode the conditional input modality. The target modality is generated while preserving Text2Earth’s inherent image generation process. 
% By training only the newly introduced module parameters, it reduced training costs while maintaining the flexibility and quality of the generated images.

\begin{figure*}
	\centering
 % \vspace{-20pt}
	\includegraphics[width=1\linewidth]{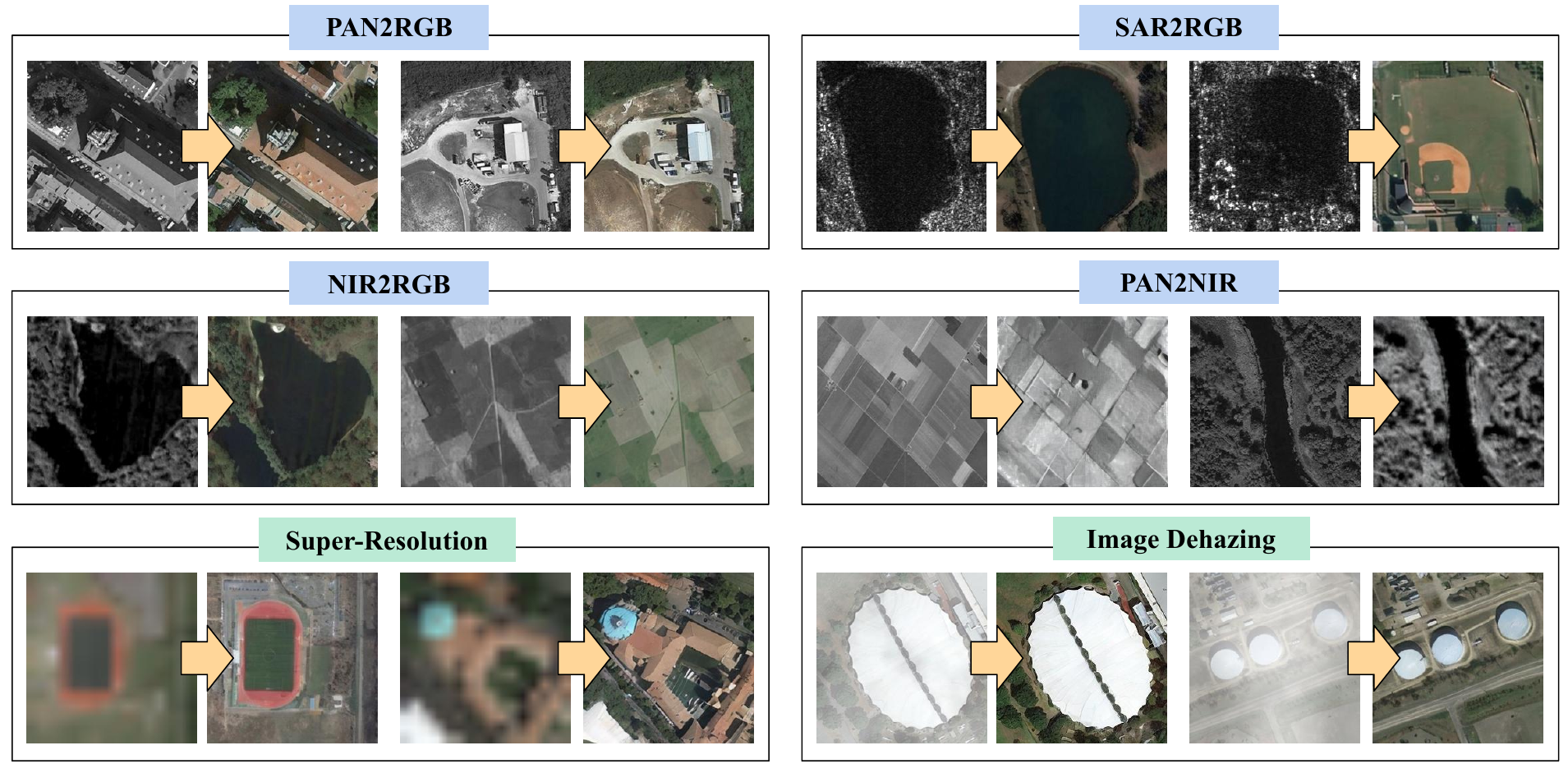}
        % \vspace{-15pt}
	\caption{Image-to-image translation, containing cross-modal translation and image enhancement, such as PAN to RGB (PAN2RGB), NIR to RGB (NIR2RGB), and PAN to SAR (PAN2SAR), low-resolution-to-high-resolution (LR2HR), and image defogging.
 }
	\label{fig:ModalTransfer}
% \vspace{-15pt}
\end{figure*}

We conducted image-to-image translation experiments using the RSICD dataset, covering tasks like PAN2RGB, NIR2RGB, and PAN2SAR. The results, as shown in Fig. \ref{fig:ModalTransfer}, demonstrate that Text2Earth effectively translates between different modalities with high fidelity. For example, in the NIR2RGB translation, the generated RGB images faithfully represent vegetation cover areas corresponding to high-intensity regions in the NIR images, adhering to the physical properties of NIR imaging. In the super-resolution task, the model exhibits remarkable detail recovery capabilities, effectively performing large-scale image super-resolution. Additionally, for the image dehazing, the model can effectively remove fog to enhance image quality. These results further validate Text2Earth’s ability to capture and transfer semantic features across different modalities, producing cross-modal images with high quality and consistency.

In summary, the experimental results above demonstrate Text2Earth’s outstanding performance in both multi-modal and cross-modal remote sensing image generation tasks. Its adaptability and extensibility as a generative foundation model make it a promising tool for a wide range of applications, including remote sensing image generation, image enhancement, and multimodal data analysis.
% We believe that Text2Earth has the potential to unlock new opportunities and breakthroughs in the remote sensing domain, driving progress across various real-world use cases.

% The above experiments demonstrate Text2Earth’s outstanding performance in multimodal and cross-modal remote sensing image generation tasks, showcasing its adaptability and extensibility as a generative foundation model. We believe that Text2Earth holds great potential for broader applications, including remote sensing image generation, enhancement, and multimodal data analysis, providing new opportunities and breakthroughs in the remote sensing domain.

\subsection{{More Applications}}
\subsubsection{\textbf{Data Augmentation}}
We explored using Text2Earth-generated synthetic images as a data augmentation tool for remote sensing scene classification. Specifically, we selected 1,027 text-image-category triplets from the RSICD dataset as training samples and generated over 20,000 synthetic images based on their textual descriptions. We then trained four widely used classification models, including VGG-19, ResNet-18, ViT-B-16, and Swin-S, under two configurations: (i) using only the original training samples and (ii) using a combination of the original samples with the synthetic images. As demonstrated in Table \ref{tab:down_classification}, all models showed consistent performance improvements when augmented with the generated images, thereby confirming that Text2Earth can serve as an effective data augmentation engine.

% {We explored the potential of using the generated remote sensing images as augmented data for the scene classification task. Specifically, we selected 1027 text-image-category triplets from the RSICD dataset as training samples. We then used our Text2Earth model to generate more than 20,000 synthetic images based on the textual descriptions of these training samples, thus creating additional ``text-synthetic\_image-category" training samples. We trained four typical classification models, including VGG-19, ResNet-18, ViT-B-16, and Swin-S, separately on (i) the original real training samples and (ii) the original real training samples combined with the generated synthetic samples. The two trained models are evaluated on the RSICD test set.} As shown in Table \ref{tab:down_classification}, the results show a consistent performance improvement across all models when incorporating the generated images, demonstrating that our Text2Earth model can effectively serve as a data augmentation engine.

\begin{table}%[htbp]%[!t]  %\small
\renewcommand{\arraystretch}{1.3}
\caption{{Accuracy of the downstream remote sensing image classification task w/ and w/o data augmentation.}}
\label{tab:down_classification}
\centering
\centering
\begin{tabular}{c|c c c c}
	\toprule%[1pt]
        Training Data & VGG-19 & ResNet-18 & ViT-B-16 & Swin-s \\
        \midrule
        w./o. Augment & 85.39 & 87.24 & 90.94 & 92.21\\
        w. Augment& 91.04 & 93.86 & 94.74 & 96.10 \\
	\bottomrule
\end{tabular}
\end{table}

\subsubsection{\textbf{Remote Sensing Vision-Language Contrastive Pretraining Foundation Model}}
We further explored the application of our Git-10M dataset to pretrain a vision-language foundation model using the contrastive learning framework. We named this model Git-RSCLIP. We then conducted zero-shot classification experiments on multiple publicly available remote sensing image classification datasets by computing the similarities of images and textualized scene category prompts to evaluate the performance of our Git-RSCLIP model. In Table \ref{tab:RSCLIP}, the experimental results demonstrate that Git-RSCLIP significantly outperforms previous remote sensing CLIP models, such as RemoteCLIP and GeoRSCLIP, confirming the effectiveness of our large-scale dataset. We have made the Git-RSCLIP model publicly available on our project page: \emph{\url{https://github.com/Chen-Yang-Liu/Text2Earth}}

\begin{table*}%[htbp]%[!t]  
\renewcommand{\arraystretch}{1.3}
\caption{{Comparison of zero-Shot classification accuracy between our model and the previous CLIP Models on multiple remote sensing scene classification datasets.}}
\label{tab:RSCLIP}
\centering
\centering
\begin{tabular}{c| cccccc|c}
	\toprule%[1pt]
        Method & OPTIMAL31 & RSC11 & RSICB128 & WHURS19 & RSSCN7 & CLRS & Average \\
        \midrule
        CLIP~\cite{CLIP} & 60.00 & 45.29 & 25.23 & 77.41 & 52.25 & 56.48 & 52.78 \\
        SkyCLIP50~\cite{wang2024skyscript} & 77.31 & 60.47 & 38.60 & 78.31 & 55.07 & 61.03 & 61.80 \\
        RemoteCLIP~\cite{liu2024remoteclip} & 81.99 & 67.05 & 34.25 & 92.54 & 51.71 & 66.04 & 65.60 \\
        GeoRSCLIP~\cite{zhang2024rs5m} & 83.33 & 67.37 & 35.48 & 89.45 & 62.54 & 69.67 & 67.97 \\
        \midrule
        Git-RSCLIP (Ours) & 95.00 & 66.96 & 52.25 & 93.93 & 63.50 & 65.18 & 72.80 \\
	\bottomrule
\end{tabular}
\end{table*}

\section{{Limitation and Discussion for Text2Earth Model}}
\begin{figure}
	\centering
 % \vspace{-20pt}
	\includegraphics[width=1\linewidth]{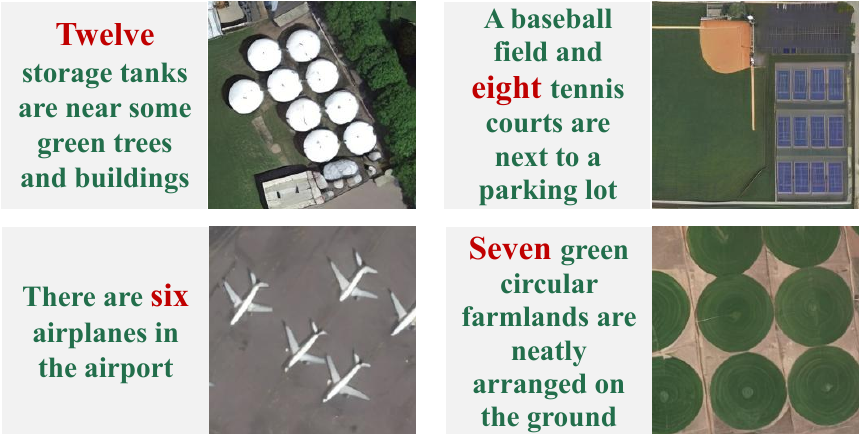}
        % \vspace{-15pt}
	\caption{{Some failure cases about inaccurate control over object quantity. These results suggest that, although Text2Earth can capture numerical cues to a certain extent, it struggles with fine-grained numerical control.}
 }
	\label{fig:limitation}
% \vspace{-15pt}
\end{figure}

% Despite the good performance of Text2Earth in large-scale text-driven remote sensing image generation, there are still limitations that deserve further discussion. 
While our Text2Earth model demonstrates robust performance in large-scale text-driven remote sensing image generation, it exhibits certain limitations that merit further discussion. One notable limitation is its inability to precisely control the number of objects specified in the textual descriptions, particularly when a large quantity is involved. For instance, as illustrated in Fig. \ref{fig:limitation}, when given the prompt ``Twelve storage tanks are near some green trees and buildings," the model generated only nine storage tanks. Similarly, the model is asked to generate seven farmlands but generates eight in the last example. These results suggest that, although Text2Earth can capture numerical cues to a certain extent, it struggles with fine-grained numerical control—a capability that is critical for accurately reflecting detailed quantitative information in generated scenes.

This limitation likely arises from the inherent challenges of aligning textual numerical information with spatial visual content during the generative process. The current model primarily focuses on learning high-level semantic relationships rather than enforcing strict quantitative constraints on object counts. To address this issue, future research could explore the integration of specialized numerical reasoning modules or enhanced conditioning strategies that explicitly account for numerical details. Additionally, incorporating more training examples with explicit numerical descriptions may further improve the model's ability to precisely control object quantity.

In summary, while Text2Earth represents a significant advancement in remote sensing image generation, addressing its limitations in object quantity control is an important avenue for future work. Improving this aspect will not only enhance the fidelity of generated images but also expand the model's applicability in real-world remote sensing applications—such as urban planning and disaster assessment—where precise quantitative control is essential.

\section{Future Work}
In this paper, we proposed a global-scale remote sensing image generation dataset and a generative foundation model based on diffusion models, Text2Earth. Through extensive experiments, we demonstrated the remarkable performance of Text2Earth across various remote sensing image generation tasks, including zero-shot image generation, image editing, unbounded scene construction, text-driven multimodal image generation, and cross-modal image generation. These achievements not only demonstrate the potential of Text2Earth in generative tasks but also open new avenues for research in the field of remote sensing image generation. Future research could focus on the following aspects.

% \begin{itemize}
% \item 
Exploring Broader Applications of Text2Earth. The advantage of Text2Earth lies in the latent knowledge it has learned from large-scale remote sensing data, especially its deep understanding of image semantics and structural information. This capability makes it well-suited not only for image generation tasks but also for promising applications such as image enhancement, object detection, and change detection. Future work could investigate how to adapt and extend Text2Earth for these domains.
% , thereby broadening its utility in remote sensing.

Developing Autoregressive Foundation Models. Autoregressive generative models, such as DALL-E \cite{ramesh2021zero_DALLE} and VAR \cite{tian2024visual_VAR} models, have shown exceptional scalability and performance in image generation, particularly under the scaling laws of large datasets. Future research could explore training autoregressive remote sensing generative foundation models with even greater representational capacity using our proposed Git-10M dataset. These models might offer advantages in terms of scalability, performance, and the ability to capture complex spatial-temporal dependencies in remote sensing data.

Building Large and Diverse Multimodal Paired Datasets.
The scale and diversity of datasets are critical drivers of advancements in generative models. While our current dataset focuses on the pairing of visible-spectrum images with text, remote sensing data contains other crucial modalities, such as SAR, NIR, and hyperspectral images. These modalities have unique physical characteristics and diverse application scenarios. Future efforts could aim to construct large-scale remote sensing datasets encompassing a broader range of paired modalities. Such datasets would not only facilitate in-depth research into cross-modal generation tasks but also advance multimodal learning in the remote sensing field.

% \end{itemize}

\section{Conclusion}
% Text2Earth excels in resolution-controllable text2image generation and demonstrates robust generalization and flexibility across multiple tasks (illustrated in Fig. \ref{fig:tasks}).
Previous remote sensing text2image generation research faces challenges in terms of dataset size and model capabilities. To this end, we present Git-10M, a global-scale remote sensing image-text pair dataset, covering diverse geographic regions globally and including rich resolution and geospatial metadata. Based on this dataset, we developed the Text2Earth foundation model, which overcomes the limitations of previous methods in terms of global-scale, multi-resolution controllable, and unbounded text2image generation. The experiments demonstrate that Text2Earth not only excels in zero-shot text2image generation but also demonstrates robust generalization and flexibility across multiple tasks such as image editing, and cross-modal translation. On the previous benchmark dataset, Text2Earth surpasses the previous models with a significant improvement of +26.23 FID and +20.95\% Zero-shot Cls-OA metric. As a generative foundation model, Text2Earth has the potential to advance a broader range of remote sensing image generation and processing tasks.

\ifCLASSOPTIONcaptionsoff
\newpage
\fi

\bibliographystyle{IEEEtran}
\bibliography{papers.bib}

\end{document}